\documentclass[11pt]{scaleai-paper}

\usepackage{makecell}
\usepackage{amsmath}
\usepackage{amssymb}
\usepackage{float}
\usepackage{booktabs}
\usepackage{tabularx}
\usepackage{subcaption}
\usepackage{natbib}
\usepackage{tikz}
\usetikzlibrary{arrows.meta, positioning, calc, shapes.geometric}
\usepackage{pifont}
\definecolor{caseGray}{HTML}{6B6963}
\definecolor{caseBlue}{HTML}{1B5CAD}
\definecolor{casePaleBlue}{HTML}{E8F0FC}
\definecolor{tableStripe}{HTML}{F5F7FA}
\definecolor{contSearchBlue}{HTML}{2A78D6}
\definecolor{passiveOrange}{HTML}{EB6834}
\newtcolorbox{turnprompt}[2]{
  enhanced, breakable,
  colback=scaleLightGray!20!white,
  colframe=#1!65!white,
  boxrule=0.55pt,
  arc=3pt,
  left=8pt, right=8pt, top=6pt, bottom=6pt,
  title={#2}, fonttitle=\bfseries\small,
  colbacktitle=#1!12!white, coltitle=black,
  attach boxed title to top left={xshift=8pt, yshift=-2.2mm},
  boxed title style={boxrule=0pt, colback=#1!12!white}
}
\usepackage{fvextra}
\usepackage{pmboxdraw}
\definecolor{noteNavy}{HTML}{1C2541}
\definecolor{notePaper}{HTML}{FFFDF7}
\definecolor{noteMark}{HTML}{F9DE82}
\tcbset{sticky/.style={
  enhanced,
  colback=notePaper, colframe=noteNavy, boxrule=0.8pt, arc=2pt,
  left=7pt, right=7pt, top=8pt, bottom=6pt,
  shadow={1.3pt}{-1.3pt}{0pt}{noteNavy},
  title={#1}, fonttitle=\sffamily\bfseries\scriptsize,
  colbacktitle=noteNavy, coltitle=white,
  attach boxed title to top left={xshift=8pt, yshift=-2.4mm},
  boxed title style={arc=3pt, boxrule=0pt}}}
\newtcolorbox{stickynote}[1]{sticky={#1}, breakable, before skip=14pt, after skip=12pt}
\newtcolorbox{stickycell}[1]{sticky={#1}}
\DefineVerbatimEnvironment{promptverb}{Verbatim}{breaklines, breakanywhere, breaksymbolleft={}, breakindent=0pt, fontsize=\fontsize{6.5}{7.9}\selectfont}
\DefineVerbatimEnvironment{promptdiff}{Verbatim}{breaklines, breakanywhere, breaksymbolleft={}, breakindent=0pt, fontsize=\fontsize{6.5}{7.9}\selectfont, commandchars=\|\[\]}

\newtcolorbox{trajectoriespan}[3][]{
  enhanced, breakable=false,
  colback=scaleLightGray!20!white,
  colframe=#2!65!white,
  boxrule=0.55pt,
  arc=3pt,
  left=6pt, right=6pt, top=4pt, bottom=4pt,
  fontupper=\footnotesize,
  title={#3}, fonttitle=\bfseries\scriptsize,
  colbacktitle=#2!12!white, coltitle=black,
  attach boxed title to top left={xshift=6pt, yshift=-2.0mm},
  boxed title style={boxrule=0pt, colback=#2!12!white},
  #1
}
\usepackage[colorlinks=true,breaklinks=true,linkcolor=scaleLink,citecolor=scaleLink,urlcolor=scaleLink]{hyperref}
\usepackage{xurl}
\usepackage{etoolbox}
\apptocmd{\thebibliography}{\raggedright}{}{}

\contact{\fontfamily{cmtt}\selectfont \{harsh.raj, vipul.gupta, david.lee, darvin.yi\}@scale.com}

\title{Root-Cause Attribution Is a Search Problem: Continual Search for Long-Horizon Agent Failures}

\author{Harsh Raj}
\author{David Lee}
\author{Anas Mahmoud}
\author{Renxiong Wang}
\author{Razvan-Gabriel Dumitru}
\author{Chenguang Wang}
\author{Tong Zhao}
\author{Yunzhong He}
\author{Darvin Yi}
\author{Vipul Gupta}
\affil{Scale AI}

\begin{document}
\maketitle

\begin{abstract}
The increasing deployment of AI agents in long-horizon tasks yields massive execution logs. Diagnosing failures within these records is crucial for reliability, as it transforms outcome-level signals into actionable interventions. The sheer scale of the data renders human review impractical, driving the need for automated root-cause attribution (RCA). However, automated RCA methods using LLMs suffer from low diagnostic accuracy, especially as execution traces grow larger. They struggle because relevant information is often sparse, distributed across distant actions, and disconnected from the visible failure, reducing root-cause attribution to a massive search problem.
Existing RCA methods typically rely on one-shot LLM judgments to diagnose failures from execution traces. While effective for shorter trajectories, these judges tend to settle on a plausible diagnosis early, leaving critical evidence in longer traces unexamined. 
We introduce Continual Search, an iterative framework that nudges the judge, over successive turns, to keep searching for unresolved diagnostic evidence.
We evaluate Continual Search across four existing RCA  benchmarks. Recognizing the lack of massive execution traces in current benchmarks, we introduce MegaRCA-Mix to evaluate RCA at scale. MegaRCA-Mix provides a challenging testbed of 50 human-annotated failure trials spanning long-horizon, execution-heavy tasks. Across multiple benchmark suites and model families, Continual Search consistently improves attribution performance. On MegaRCA-Mix, for example, it improves Opus-4.8's F1 score by 29\%, from $0.471$ to $0.608$. More interestingly, within the same model family, lower-tier models can even surpass their higher-tier counterparts, demonstrating that effective search supersedes raw model scale.
\end{abstract}

\section{Introduction}
\label{sec:intro}

\begin{figure}[H]
\centering
\includegraphics[width=\textwidth]{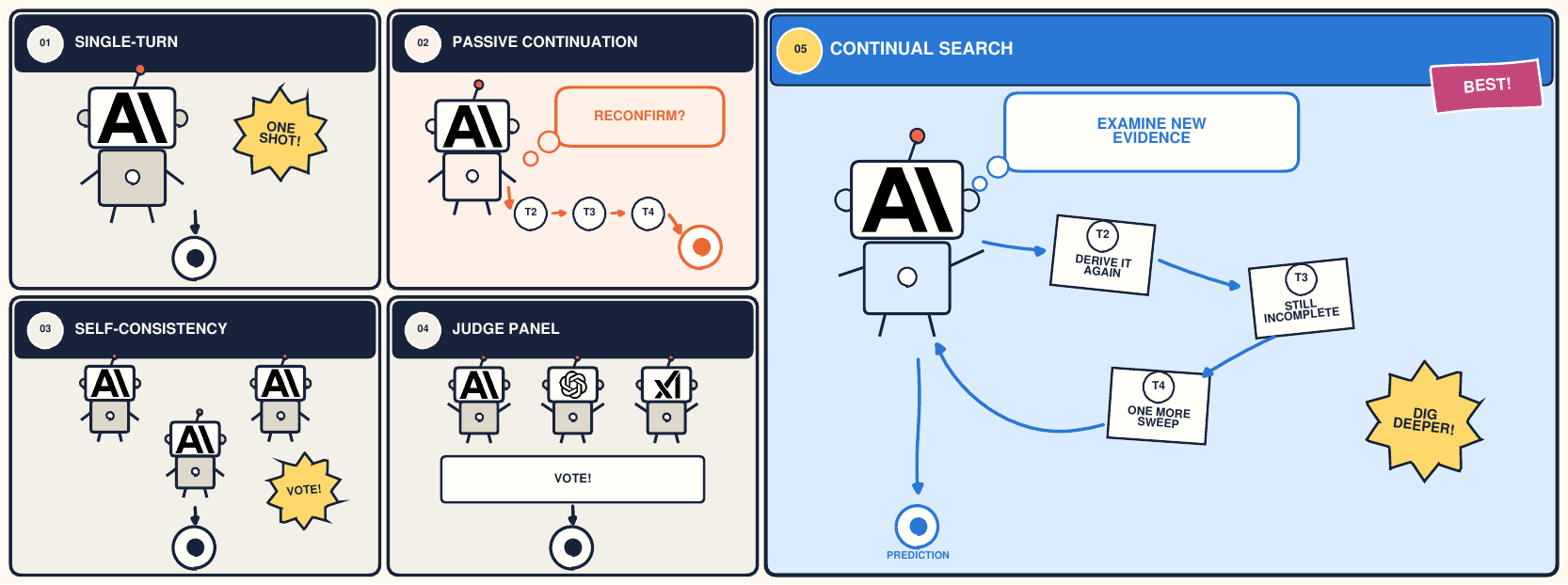}
\caption{\textbf{Comparison of root-cause attribution strategies.} Standard baselines like one-shot judging, self-consistency, and a heterogeneous judge panel perform a single pass over the evidence. \textit{Passive Continuation} prompts the judge to repeatedly reconsider its diagnosis within an ongoing session but does not mandate further exploration. 
Conversely, \textit{Continual Search} asks the judge to challenge its standing answer at each successive turn and search beyond the evidence that supported it.}
% \nas{why can't we just use \textbf{explore unobserved evidence} on its own to force the model to explore what is hasn't explored before. This seems to be a general prompt (i.e., independent of benchmark). The judge looks at what was previously explored and tries to expand the search space. The point here is: why do you think that's not enough and that we have to point the judge to which artifacts to explore.}, \harsh{unrealiable label, derive it again or treat everything found so far as still incomplete}\harsh{trail to be first and megarca-mix to the last} 
\label{fig:intro-length}
\end{figure}

As AI agents are deployed on increasingly long-horizon tasks, agent reliability has emerged as a primary bottleneck. While outcome-level failure signals indicate that an execution failed, they offer minimal insight into the underlying failure mechanisms and where to intervene. Root-cause attribution (RCA) turns this outcome-level signal into actionable diagnostic feedback~\citep{wang2026surveytrajectory, chang2026strace}.  It helps identify the root cause—the first divergence from a correct execution. The fault is then assigned to the responsible component, such as the model, harness, environment, or grader~\citep{raj2026modelharness}. This distinction directs intervention to the appropriate part of the agentic system: model failures can inform post-training objectives, harness failures can guide harness design, and environment or grader issues can trigger benchmark repair. Such diagnosis becomes increasingly important as agentic systems move toward self-improvement: recent systems learn from failed trajectories, iteratively refine their behavior, and even modify components of their own agent architecture \citep{yuan2025agentr, sun2026learningfailure, zhang2025darwin}. Root-cause attribution therefore provides a feedback layer between failed executions and the continual improvement of the broader agentic system.

RCA becomes especially important when agent failures have security or safety consequences. In a recent Hugging Face security incident \citep{openai2026huggingface}, an autonomous agent driven by OpenAI models first exploited vulnerabilities in OpenAI's evaluation environment and then gained access to Hugging Face's production infrastructure. Understanding what happened required reconstructing a long sequence of actions across both the environments. Investigators recovered and analyzed more than 70,000 agent messages and files to reconstruct the intrusion timeline \citep{huggingface2026incident}. Moreover, reconstructing the intrusion required repeated passes over the execution record, with each pass surfacing evidence earlier readings had missed.

% \harsh{1. use 70k, 2. Even in the huggingface incident they went through the traces again and again and thats why we also test this multi-turn judge to ensure reliability, quota this: https://openai.com/index/hugging-face-incident-and-the-road-ahead/. Remove the last line of the para above and add this line there. Basically that addresses Nas' comments on why multi-turn is important because openai looked at their traces again and again}

The forensic burden in such incidents becomes harder to manage as agent systems operate over longer horizons and involve an increasing number of interactions. Long-running agents accumulate actions, tool calls, environment states, retries, and intermediate outputs over time, while multi-agent systems can distribute relevant evidence across many interacting agents. As these executions grow, their execution logs can reach up to millions of tokens, and the evidence needed to explain a failure may be sparse, appear much earlier than the final outcome, or be distributed across distant parts of the execution-logs. 
Root-cause attribution is therefore better characterized as search problem: the attribution system must search through a large evidence space, test multiple plausible explanations, and rule them out as contradictory evidence is found until it identifies the underlying root cause. At production scale, reliable attribution becomes critical, since misattributions incur substantial cost and can even trigger unnecessary interventions.
Existing automated RCA methods typically use LLM judges to inspect execution traces and produce an attribution through a single rubric-guided pass \citep{deshpande2025trail,barke2026agentrx,wang2026drift}. Agent-as-a-Judge frameworks extend this approach by enabling judges to inspect intermediate execution evidence through tool use \citep{zhuge2024agentjudge}. To further improve both the performance and reliability of these judge-based systems, strategies like self-consistency or heterogeneous judge panels are commonly used~\citep{wang2023selfconsistency,verga2024poll}.

% \nas{Another way to demonstrate that RCA is challenging due to the large search space is to summarize (i.e., using an LLM) the trajectories (actions + environment observations) and then show that the classification performance for the summarized trajectories is higher than raw trajectories. The point here is to demonstrate that the same failure is much easier to classify/find when the trajectory is shorter. I understand that this doesn't take into account the workspace files for a task but it would still show the challenge of RCA as a function of trajectory length for a fixed task}. 

% \vipul{we can also mention something along the lines that: this becomes really critical when we are looking are millions of traces to identify the root cause. The reliability of the judge is very critical both from time and cost perspective}

However, as execution logs become larger and more distributed, LLMs tend to settle on a plausible-looking failure as the root cause before fully exploring the evidence space~\citep{mehta2026agentscommitsoondiagnosing}. To address this problem, we introduce Continual Search, an iterative framework that nudges the judge, over successive turns, to keep exploring unresolved or previously unexamined evidence. We evaluate it on four existing RCA benchmarks, TRAIL~\citep{deshpande2025trail}, TELBench~\citep{wang2026drift}, AgentRx~\citep{barke2026agentrx}, and Who\&When~\citep{zhang2025whowhen}, keeping each benchmark's native scoring and localization schema. On the two benchmarks with the longest execution-logs, TRAIL and TELBench, Continual Search consistently improves attribution over passive reconsideration. On TRAIL, for example, it raises Opus-4.8's Weighted F1 from $0.482$ to $0.543$ over four turns, compared with $0.488$ under Passive Continuation.  It also outperforms other judge baselines such as self-consistency~\citep{wang2023selfconsistency} and heterogeneous judge panels~\citep{verga2024poll,kohli2026judgepanels}~(Table~\ref{tab:ensemble-judges}). These gains coincide with the judge reading new evidence. Over the same four turns, the deduplicated evidence it reads grows from $20$K to $24$K tokens under Continual Search, but only to $21$K under Passive Continuation (Figure~\ref{fig:dedup-observation}). On the shorter AgentRx and Who\&When trajectories, however, the initial attribution already covers most of the available evidence. With little left to search, additional turns no longer meaningfully expand the evidence base. Instead, they place pressure on the judge to revisit its existing prediction, often flipping it toward an incorrect label~\citep{laban2024surechallengingllmsleads,zhao2026jaggedjudges}.

To evaluate RCA on long-horizon executions, we introduce MegaRCA-Mix, a collection of 50 failure trials drawn from Harbor Index~\citep{harborindex2026} execution logs, each human-annotated with its root cause under the interaction-centric taxonomy of \citet{raj2026modelharness}. With a median execution size of 286K tokens, MegaRCA-Mix is substantially larger than existing RCA benchmarks (Table~\ref{tab:datasets}). Beyond the agent trajectory, it provides the full evaluation record, including trial configurations, agent-side logs, verifier outputs, and sandbox artifacts, reflecting how agents are now evaluated in execution-heavy, sandbox-based environments~\citep{harborframework2026,merrill2026terminalbenchbenchmarkingagentshard}. On MegaRCA-Mix, Continual Search improves Opus-4.8's F1 from $0.471$ to $0.608$, compared with $0.479$ under Passive Continuation. 

\paragraph{Contributions.}
\begin{itemize}

\item \textbf{Continual Search.}
We introduce Continual Search, an iterative framework that nudges an agentic LLM judge to keep searching for  unresolved or previously unexamined diagnostic evidence. On benchmarks with large execution logs, Continual Search yields monotonic attribution gains across turns, matching or outperforming every attribution method we evaluate. 

% \harsh{outperforms all other tools to get the RCA} judge baselines.
% \vipul{We should say our motivation in a different way something along the lines that we are nudging the judge to find the unexamined evidence in the trace and improve its final diagnosis -- change intro and abstract as it will clear Nas' comment}

\item \textbf{MegaRCA-Mix.}
We introduce MegaRCA-Mix to extend RCA evaluation to substantially larger evidence spaces than those covered by existing benchmarks. MegaRCA-Mix contains 50 failure trials with human-annotated root causes and a median execution-log size of 286K tokens. Beyond the agent trajectory, each trial includes the full set of execution artifacts, reflecting how agents are evaluated in sandbox-based environments.

\item \textbf{Frontier models as judges do not necessarily improve RCA.}
Experiments with five judge models and multiple reasoning-effort settings further show that long-horizon RCA is primarily search-limited. At the same reasoning-effort setting, Sonnet-5 and Fable-5 reach comparable performance using Continual Search.

\end{itemize}

\section{Related Work}
\label{sec:related}

\paragraph{Root-cause attribution for agent failures.}
To provide actionable diagnostic feedback for agentic failures, recent studies investigate where and why these executions fail. \citet{raj2026modelharness} formulate failure attribution in terms of the interaction between two components in an agentic system and the fault side responsible for a failure. Who\&When \citep{zhang2025whowhen} identifies the responsible agent and decisive error step in multi-agent executions, while AgentRx \citep{barke2026agentrx} localizes critical steps and assigns root-cause categories from execution trajectories. Who\&When Pro \citep{whowhenpro2026} substantially expands the breadth of this evaluation, with more than 12K labeled trajectories across agent frameworks, domains, and modalities. Yet, these datasets still largely operate at a scale where the full trajectory can be easily ingested and reasoned over by modern LLMs with context windows of over a hundred thousand tokens~\citep{qwen2025qwen25technicalreport}. For example, Who\&When Pro averages just 7.5 steps per trajectory, with its longest reported analysis bin starting above 12K tokens. Similarly, AgentRx trajectories average between 4.9K and 16.5K tokens across its three domains.

\paragraph{Evidence search in long-horizon trajectories.} Root-cause attribution becomes more challenging on frontier benchmarks, where contemporary agents can produce execution logs containing hundreds of thousands of tokens \citep{huang2026deepswemeasuringfrontiercoding,desai2026swemarathonagentsautonomouslycomplete}. Larger context windows do not necessarily make these logs easier to analyze. Relevant information may be diluted by accumulated context \citep{activecontext2026}, distributed across distant tool interactions \citep{acc2026}, or lose influence as the execution progresses \citep{proactivememory2026,longhorizonharness2026}. Root-cause attribution therefore requires the judge to search the execution log for evidence relevant to the diagnosis. Results and analysis from TRAIL \citep{deshpande2025trail} and TrajDebug \citep{trajdebug2026} further demonstrate the difficulty of localizing errors when such evidence is scattered across an execution.

\paragraph{Agent-as-a-Judge.}
Agent-as-a-Judge systems employ tool-enabled evaluators that can inspect and gather evidence before issuing a verdict \citep{zhuge2024agentjudge}. This setup is particularly well-suited for long execution artifacts, where relevant evidence may be distributed across the trajectory and must be retrieved selectively during analysis. More broadly, iterative prompting and self-critique methods have demonstrated that additional rounds of critique can improve an initial prediction. For instance, Constitutional AI, Self-Refine, and Reflexion use critique and revision to refine baseline generations \citep{bai2022constitutional,madaan2023selfrefine,shinn2023reflexion}. Multi-agent debate similarly relies on several rounds of interaction before reaching a consensus \citep{du2024debate}.
Additional interaction, however, does not necessarily improve a judgment. Repeated interaction can introduce pressure on the model's judgment. Instruction-tuned models may defer to conversational pressure even when their initial answer is correct \citep{perez2022sycophancy,sharma2023sycophancy}. Recent work further shows that repeated pressure can flip a judge's verdict \citep{zhao2026jaggedjudges,dutta2026stability}.
\citet{kim2025challenging} find that judges are more susceptible to a
conflicting argument when it is introduced gradually across multiple turns.

\section{Method}
\label{sec:method}

\subsection{Continual Search} \label{sec:continual-search}
\label{sec:continuation-strategies}
\begin{figure}[htbp]
\centering
\includegraphics[width=\textwidth]{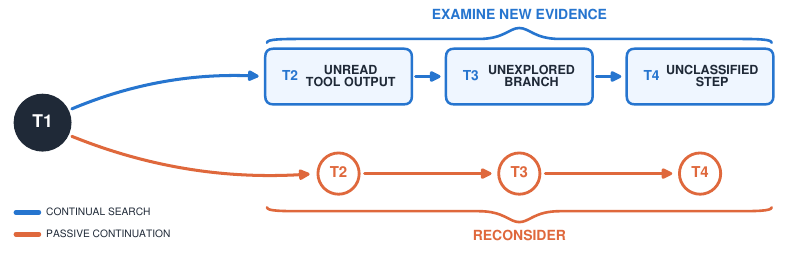}
\caption{\textit{Continual Search} and \textit{Passive Continuation} start from the same turn 1 (T1) attribution. \textit{Passive Continuation} only asks the judge to reconsider and reverify its answer, whereas \textit{Continual Search} prompts it to challenge its current diagnosis by searching  for evidence it has not yet examined.}
\label{fig:continuation}
\end{figure}
We study root-cause attribution as a search over the evidence contained in an agent's execution record. To extend this search beyond the initial attribution, we use an iterative multi-turn setting in which the judge is prompted to continue its analysis. For each sample, we first run the benchmark's native single-turn RCA method, typically a rubric-guided prompt, using an agentic judge. We then branch the resulting session into two continuation conditions (Figure~\ref{fig:continuation}). Both branches begin with the same judge configuration, evaluation record, task instructions, and turn~1 prediction. The instruction provided in subsequent turns is the only variable that differs between the two conditions.

As a control, \textit{Passive Continuation} asks the judge to reconsider and reverify its current conclusion without asking it to examine additional evidence. In contrast, \textit{Continual Search} asks the judge to expand its analysis. Prior work shows that language models frequently resist revising conclusions to which they have already committed \citep{tsui2025selfcorrection}. Continual Search addresses this tendency by nudging the judge to challenge its standing conclusion and to examine unresolved or previously unexamined evidence. To keep the method broadly applicable, we do not write benchmark-specific prompts. Instead, a decision tree assembles the prompts for turns 2--4 from a single fixed set of generic sentences, written once without reference to any benchmark. The tree chooses among these sentences based on two properties of each benchmark, listed in Table~\ref{tab:datasets}: how its evidence is stored and how its answer is scored. Turn 1 retains each benchmark's original prompt, and Passive Continuation uses one fixed sentence throughout. Appendix~\ref{sec:appendix-prompts} describes the tree in full and reproduces every prompt verbatim.

\subsection{Evaluation Setting} \label{sec:evaluation-setting}
We evaluate Continual Search across five distinct root-cause attribution settings: MegaRCA-Mix, TRAIL~\citep{deshpande2025trail}, TELBench~\citep{wang2026drift}, AgentRx~\citep{barke2026agentrx}, and the Algorithm-Generated subset of Who\&When~\citep{zhang2025whowhen}. These benchmarks include diverse failure targets and scoring procedures, spanning interaction-centric failure taxonomy classification, fine-grained error span and step localization, and multi-agent attribution. By retaining each benchmark's native task formulation and evaluation metrics, we assess Continual Search independently of any specific RCA schema. Table~\ref{tab:datasets} summarizes the attribution target, dataset size, and execution scale for each setting. Each benchmark is scored with its native evaluation metric, and the exact formulations are given in Appendix~\ref{sec:appendix-metrics}.

\begin{table}[t]
\centering
\footnotesize
\setlength{\tabcolsep}{4pt}
\renewcommand{\arraystretch}{1.08}
\rowcolors{2}{tableStripe}{white}
\begin{tabularx}{\textwidth}{
    @{}
    >{\raggedright\arraybackslash}p{0.14\textwidth}
    >{\raggedright\arraybackslash}X
    >{\centering\arraybackslash}p{0.035\textwidth}
    >{\centering\arraybackslash}p{0.075\textwidth}
    >{\centering\arraybackslash}p{0.075\textwidth}
    @{}
}
\toprule
Dataset &
Description &
$n$ &
\makecell[c]{Median\\tokens} &
\makecell[c]{Median\\size} \\
\midrule

MegaRCA-Mix &
First consequential error in Harbor Index evaluation records
\citep{harborindex2026}, labeled with the interaction-centric taxonomy
\citep{raj2026modelharness}. The mixture spans GAIA \citep{mialon2023gaia}, SWE-Bench
\citep{swebench_verified,swebenchpro2025}, HLE \citep{phan2025hle}, OpenRCA
\citep{openrca},
and 15 additional benchmarks. &
50 & 286K & 1.05 MiB \\

TRAIL \citep{deshpande2025trail} &
Error location and category in GAIA \citep{mialon2023gaia} and
SWE-Bench \citep{swebench_verified} trajectories &
148 & 100K & 430 KiB \\

TELBench \citep{wang2026drift} &
Earliest harmful error span in deep-research trajectories from GAIA
\citep{mialon2023gaia}, BrowseComp \citep{wei2025browsecomp}, and
xbench \citep{chen2025xbench} &
33 & 39K & 144 KiB \\

AgentRx \citep{barke2026agentrx} &
Critical failure step in retail $\tau$-bench
\citep{yao2024taubench} trajectories &
29 & 7.2K & 23.9 KiB \\

Who\&When \citep{zhang2025whowhen} &
Responsible agent and decisive error step in multi-agent trajectories over
GAIA \citep{mialon2023gaia} and AssistantBench
\citep{yoran2024assistantbench} &
126 & 2.4K & 8.9 KiB \\

\bottomrule
\end{tabularx}
\caption{Overview of the root-cause attribution benchmarks. The table details each benchmark's attribution target, number of annotated failure trials ($n$), and the median token count and raw file size of its execution evidence. Native evaluation metrics are defined in Appendix~\ref{sec:appendix-metrics}.}
\label{tab:datasets}
\end{table}
MegaRCA-Mix consists of 50 failed execution trials drawn from Harbor Index~\citep{harborindex2026}, generated by running Claude Code~\citep{claudecode2026} with Opus-4.8 and GPT-5.5 using the Harbor evaluation framework~\citep{harborframework2026}. Each trial is human annotated with its first consequential error according to the interaction-centric taxonomy of \citet{raj2026modelharness}. Harbor Index spans 29 benchmarks across seven domains—including software engineering, scientific research, mathematics, data analytics, and security. Consequently, MegaRCA-Mix captures a diverse set of audited agent failures rather than narrow, task-specific errors. For each trial, the judge has access to the complete evaluation record, including the raw execution trajectory, trial configuration, agent-side logs, verifier outputs, and sandbox artifacts.

Beyond their distinct attribution targets, these five settings span a wide range of execution lengths (Table~\ref{tab:datasets}). MegaRCA-Mix features a median size of 286K tokens, followed by TRAIL at 100K tokens and TELBench at 39K tokens. In contrast, AgentRx and Who\&When are substantially shorter, with median lengths of 7.2K and 2.4K tokens, respectively. This spectrum allows us to evaluate how the benefits of Continual Search scale as the evidence space expands. The scale of these evaluation records also informs how evidence is presented to the judge. Inlining massive execution traces directly into the context window risks information saturation and context limit exhaustion~\citep{du2025contextlengthhurtsllm}. We therefore employ tool-enabled agentic judges that selectively inspect relevant artifacts on demand. We implement our judges using the Claude Agent SDK~\citep{anthropic2026claudeagentsdk}. Each judge is granted read-only access to the evaluation record and gathers evidence through explicit tool interactions. Detailed judge configurations, tool interfaces, and verbatim continuation prompts for all benchmarks are provided in Appendices~\ref{sec:appendix-judge-config} and~\ref{sec:appendix-prompts}.
\section{Results}
\label{sec:results}

\subsection{Continual Search Improves Long-Horizon Attribution}
\label{sec:continuation-results}

Continual Search consistently improves attribution on MegaRCA-Mix, TRAIL, and
TELBench, the three settings with the largest execution records in our
evaluation. Table~\ref{tab:long-trajectory-performance} compares the initial single-turn attribution against the Turn 4 results for both Passive Continuation and Continual Search using GPT-5.5 and Opus-4.8.

\begin{table}[H]
\centering
\small
\rowcolors{3}{white}{tableStripe}
\begin{tabular}{@{}l cccccc@{}}
\toprule
& \multicolumn{3}{c}{GPT-5.5}
& \multicolumn{3}{c}{Opus-4.8} \\
\cmidrule(lr){2-4}\cmidrule(lr){5-7}
Dataset & Single-turn & Passive & Continual Search
& Single-turn & Passive & Continual Search \\
\midrule
MegaRCA-Mix & 0.356 & \textbf{0.569} & \textbf{0.569}
            & 0.471 & 0.479 & \textbf{0.608} \\
TRAIL (Joint Acc.)  & 0.143 & 0.170 & \textbf{0.207}
                    & 0.133 & 0.135 & \textbf{0.152} \\
TRAIL (Weighted F1) & 0.429 & 0.459 & \textbf{0.494}
                    & 0.482 & 0.488 & \textbf{0.543} \\
TELBench    & 0.182 & 0.212 & \textbf{0.242}
            & 0.091 & 0.091 & \textbf{0.152} \\
\bottomrule
\end{tabular}
\caption{Attribution performance on MegaRCA-Mix, TRAIL, and TELBench after four continuation turns. Each benchmark is evaluated using its native metric, with TRAIL reporting both Joint Accuracy and Weighted F1. The highest score for each judge is bolded.}
\label{tab:long-trajectory-performance}
\end{table}

Across all three datasets, Continual Search yields the highest final attribution scores for both judge models, except GPT-5.5 on MegaRCA-Mix, where it ties Passive Continuation. The performance advantage is particularly pronounced for Opus-4.8 on MegaRCA-Mix, which improves from an initial $0.471$ F1 score to $0.608$ under Continual Search, compared to just $0.479$ under Passive Continuation. TRAIL and TELBench show a similar overall pattern. On TRAIL, the consistent superiority of Continual Search holds under both Joint Accuracy and Weighted F1, the latter of which strictly penalizes the prediction of ungrounded error categories. Across these long-horizon settings, Continual Search yields steady improvements that fundamentally outpace both single-turn attribution and Passive Continuation.

\begin{figure}[H]
\centering
\begin{minipage}{0.88\textwidth}
\raggedright\footnotesize

(a) TRAIL
\end{minipage}
\par\vspace{2pt}
\includegraphics[width=0.88\textwidth]{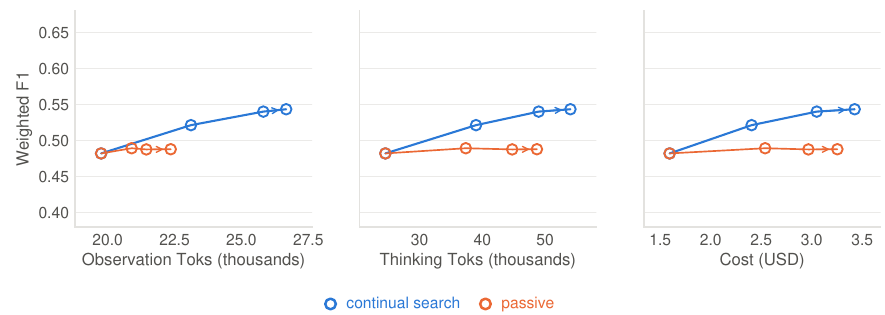}
\par\vspace{3pt}

\begin{minipage}{0.88\textwidth}
\raggedright\footnotesize
(b) TELBench
\end{minipage}
\par\vspace{2pt}
\includegraphics[width=0.88\textwidth]{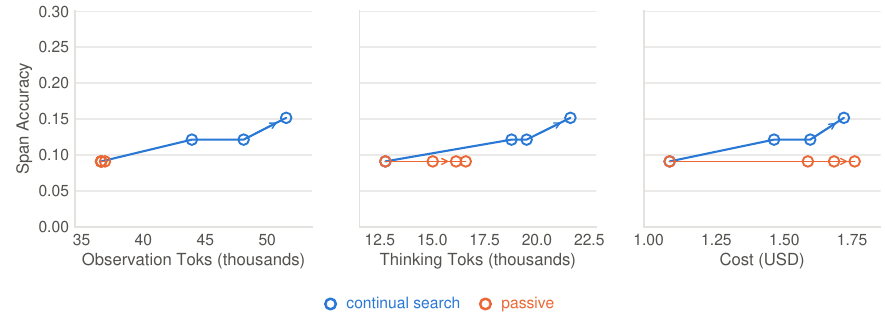}
\par\vspace{3pt}

\begin{minipage}{0.88\textwidth}
\raggedright\footnotesize
(c) MegaRCA-Mix
\end{minipage}
\par\vspace{2pt}
\includegraphics[width=0.88\textwidth]{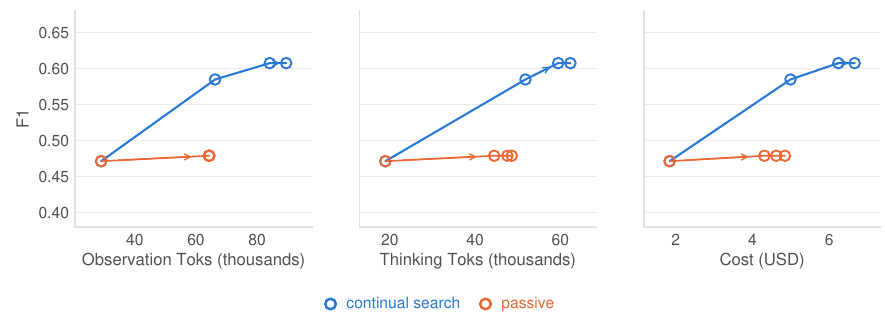}

\caption{Opus-4.8 attribution performance against cumulative observation tokens, reasoning tokens, and cost across four continuation turns. Cumulative observation tokens represent the total token ingested across all read operations executed by the agentic judge until that turn. MegaRCA-Mix is scored by F1, TRAIL by Weighted F1, and TELBench by Span Accuracy.}
\label{fig:long-trajectory-gpt55}
\end{figure}

Figure~\ref{fig:long-trajectory-gpt55} illustrates how this performance gap develops across successive turns for Opus-4.8. To isolate the mechanism driving these gains, we track cumulative observation tokens, reasoning tokens, and inference cost alongside attribution accuracy. Monitoring reasoning tokens specifically confirms that the judge actively processes newly retrieved evidence, rather than merely accumulating context under prompting pressure. As demonstrated across MegaRCA-Mix, TELBench, and TRAIL, Continual Search persistently expands both its evidence base (observation tokens) and its cognitive effort (thinking tokens) as the analysis proceeds. In contrast, the Passive Continuation baseline saturates after the initial attribution, exhibiting limited subsequent evidence acquisition. This active exploration framework effectively mitigates premature commitment in LLM agents, preventing the judge from settling on an early, plausible interpretation while ignoring unread downstream evidence \citep{mehta2026agentscommitsoondiagnosing}.

\subsection{Varying Model Scale and Reasoning Effort}
\label{sec:search-vs-capability}

We next examine how Continual Search interacts with baseline judge capability and inference-time reasoning effort. Holding the TRAIL benchmark fixed, we evaluate performance across two dimensions. First, we vary the underlying judge by comparing GLM-4.7, Sonnet-5, Opus-4.8, and Fable-5. Second, we modulate the per-turn reasoning budget across low, high, and max settings.

\begin{figure}[H]
\centering
\includegraphics[width=0.5\textwidth]{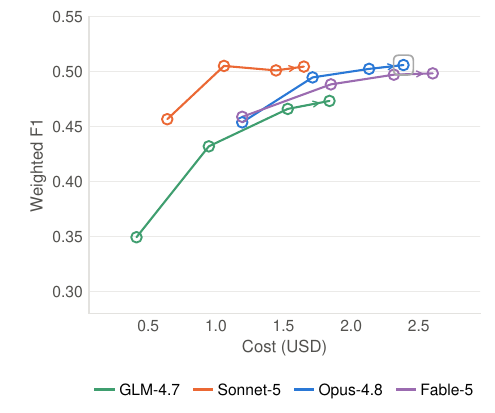}
\caption{\textbf{Impact of model tier on attribution performance.} Weighted F1 across turns 1--4 of Continual Search on TRAIL plotted against cumulative per-trajectory inference cost for GLM-4.7, Sonnet-5, Opus-4.8, and Fable-5 at default reasoning settings. Cost shows the mean price, and the boxed value marks the best Turn 4 score.}
\label{fig:model-effort}
\end{figure}

Across model tiers, Continual Search improves every judge over its turn-1 attribution, as shown in Figure~\ref{fig:model-effort}. Sonnet-5 reaches a turn 4 Weighted F1 of $0.505$, comparable to the $0.498$ of Fable-5, at about $64\%$ of its cost, and Opus-4.8 does slightly better than Fable-5 at a somewhat lower cost. Scores rise from one turn to the next with one exception, a very slight dip for Sonnet-5 from $0.505$ at turn 2 to $0.501$ at turn 3, after which it recovers. These results indicate that lower-tier models equipped with Continual Search can match or exceed frontier models operating under substantially larger compute budgets.

\begin{figure}[H]
\centering
\includegraphics[width=0.77\textwidth]{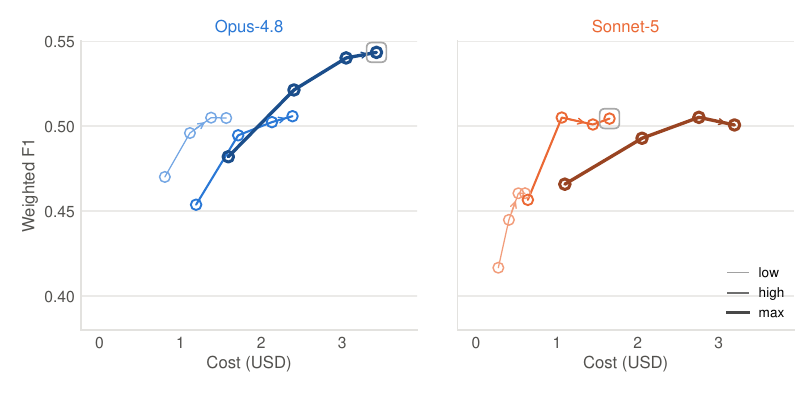}
\caption{\textbf{Impact of reasoning effort on attribution performance.} Weighted F1 across turns 1--4 of Continual Search on TRAIL plotted against cumulative per-trajectory inference cost, ablating low, high, and max reasoning settings for Opus-4.8 (left) and Sonnet-5 (right). Cost shows the mean price, and the box in each panel marks that model's best Turn 4 score.}
\label{fig:reasoning-effort}
\end{figure}

Reasoning effort ablation in Figure~\ref{fig:reasoning-effort} shows no consistent pattern. Raising effort helps Opus-4.8 only at max, which reaches $0.543$ against about $0.505$ for low and high. For Sonnet-5, low effort is worst and max is slightly below high. What is consistent is the trend across turns. Continual Search raises Weighted F1 from turn 1 to turn 4 in every configuration. Neither the model nor the reasoning effort changes this pattern, which suggests that additional evidence search is a more dependable source of improvement than either. Once a judge meets the baseline competence required to evaluate retrieved logs, inference compute is far more effectively spent on expanding evidence search across turns than on scaling per-turn reasoning effort.
%TODO: TRAIL labels incorrect in limitation

\subsection{Limited Benefit on Short Execution trajectories}
\label{sec:continual-search-limits}

The benefits of Continual Search depend fundamentally on the availability of unread evidence. AgentRx and Who\&When operate at a vastly different scale than the other three long-horizon benchmarks. Their median trajectory lengths are just 7.2K and 2.4K tokens, respectively, compared to the 39K–286K token range of MegaRCA-Mix, TRAIL, and TELBench. Given these highly constrained evidence spaces, the initial turn of the judge typically processes the entire execution log in a single pass, leaving almost no unexamined material to recover during later turns.

\begin{table}[htbp]
\centering
\small
\rowcolors{3}{white}{tableStripe}
\begin{tabular}{@{}l cccccc@{}}
\toprule
& \multicolumn{3}{c}{GPT-5.5}
& \multicolumn{3}{c}{Opus-4.8} \\
\cmidrule(lr){2-4}\cmidrule(lr){5-7}
Dataset & Single-turn & Passive & Continual Search
& Single-turn & Passive & Continual Search \\
\midrule
AgentRx   & 0.310 & \textbf{0.345} & 0.241
          & \textbf{0.414} & 0.379 & 0.379 \\
Who\&When & \textbf{0.516} & 0.492 & 0.437
          & \textbf{0.373} & 0.349 & 0.349 \\
\bottomrule
\end{tabular}
\caption{Attribution performance on AgentRx and Who\&When after four
continuation turns. In contrast to MegaRCA-Mix, TRAIL, and TELBench,
Continual Search does not consistently improve over the single-turn
attribution.}
\label{tab:short-trajectory-performance}
\end{table}

As shown in Table~\ref{tab:short-trajectory-performance}, Continual Search no longer produces consistent gains in this regime. Under GPT-5.5, performance actually degrades: AgentRx falls from a $0.310$ initial score to $0.241$, and Who\&When falls from $0.516$ to $0.437$. Opus-4.8 also regresses on both, from $0.414$ to $0.379$ on AgentRx and from $0.373$ to $0.349$ on Who\&When.

This degradation aligns with established findings regarding epistemic instability in LLM judges. Prior work demonstrates that sustained prompting pressure can force a judge to flip its verdict, frequently causing the model to drift away from the ground truth \citep{laban2024surechallengingllmsleads,zhao2026jaggedjudges}. When the initial single-turn attribution already covers the vast majority of the available evidence, Continual Search prompts act merely as conversational pressure rather than as a mechanism for meaningful discovery. The stark contrast between these short-trajectory results and the gains observed on MegaRCA-Mix, TRAIL, and TELBench confirms that Continual Search is effective precisely when further search can recover critical evidence omitted from the initial analysis.

\section{Ablations} \label{sec:ablations}

\subsection{Continual Search Outperforms Independent Resampling} We evaluate Continual Search against standard ensembling baselines to determine if independent resampling can replicate the benefits of sequential exploration. A one-shot judge produces a single baseline attribution. Self-consistency draws four independent attributions from the same model and aggregates them via majority vote \citep{wang2023selfconsistency}. A heterogeneous judge panel similarly aggregates predictions, but across four distinct models to mitigate individual judge biases \citep{verga2024poll,kohli2026judgepanels}. Unlike these resampling strategies, which perform multiple independent evaluations of the execution trace, Continual Search carries the diagnostic state forward across turns. It explicitly compels the judge to explore further unexamined evidence and critically challenge its standing attribution. Table~\ref{tab:ensemble-judges} reports the results of this comparison on the TRAIL dataset.
\begin{table}[htbp] \centering \small \rowcolors{3}{white}{tableStripe} \begin{tabular}{@{}l cc cc@{}} \toprule
& \multicolumn{2}{c}{Weighted F1} & \multicolumn{2}{c}{Cost \$} \\
\cmidrule(lr){2-3}\cmidrule(lr){4-5}
Method & Opus-4.8 & GPT-5.5 & Opus-4.8 & GPT-5.5 \\ \midrule
Single-turn & 0.482 & 0.429 & 1.59 & 2.22 \\
Self-consistency & 0.430 & 0.383 & 6.52 & 8.65 \\
Judge panel & 0.431 & 0.431 & 4.73 & 4.73 \\
Continual Search & \textbf{0.543} & \textbf{0.494} & 3.43 & 5.83 \\ \bottomrule \end{tabular}
\caption{Comparison of Continual Search against standard judge baselines on TRAIL across pooled GAIA and SWE splits, using Opus-4.8 and GPT-5.5 as underlying judge models. Self-consistency uses four samples per judge, while the judge panel combines GPT-5.5, Opus-4.7, Opus-4.8, and Grok-4.6. Cost reflects the mean price per trajectory at published API rates.} \label{tab:ensemble-judges} \end{table}

Continual Search consistently outperforms both independent resampling strategies for both judge models. With Opus-4.8, Continual Search increases Weighted F1 from 0.430 under self-consistency to 0.543. GPT-5.5 exhibits a similar leap, rising from 0.383 to 0.494. In fact, self-consistency fails to improve upon the initial one-shot attribution, regressing from 0.482 to 0.430 for Opus-4.8 and from 0.429 to 0.383 for GPT-5.5. While the heterogeneous judge panel reaches 0.431 Weighted F1 with both judges, it remains substantially below Continual Search.

These results also clarify whether the observed gains stem from sequential evidence acquisition or merely the application of additional test-time compute. As reported in Table~\ref{tab:ensemble-judges}, self-consistency consumes a larger inference budget than Continual Search for both judges yet still underperforms the initial one-shot attribution it aggregates. Although the cost of the heterogeneous judge panel is confounded by varying API rates across four distinct providers, the controlled self-consistency baseline confirms that scaling independent model calls does not recover the missing evidence required for long-horizon RCA.

\subsection{Effect of Execution-Log Size}
\label{sec:t4-by-length}

Continual Search improves attribution across the three benchmarks with the largest execution records, but aggregate scores do not show how these gains vary with execution-log size within each benchmark. We therefore sort the matched Opus-4.8 examples by raw log size, divide each benchmark into three equal-count bins, and recompute the benchmark’s native metric within each bin. We compare the initial attribution with the final Continual Search attribution.

\begin{figure}[H]
\centering
\includegraphics[width=1.03\textwidth]{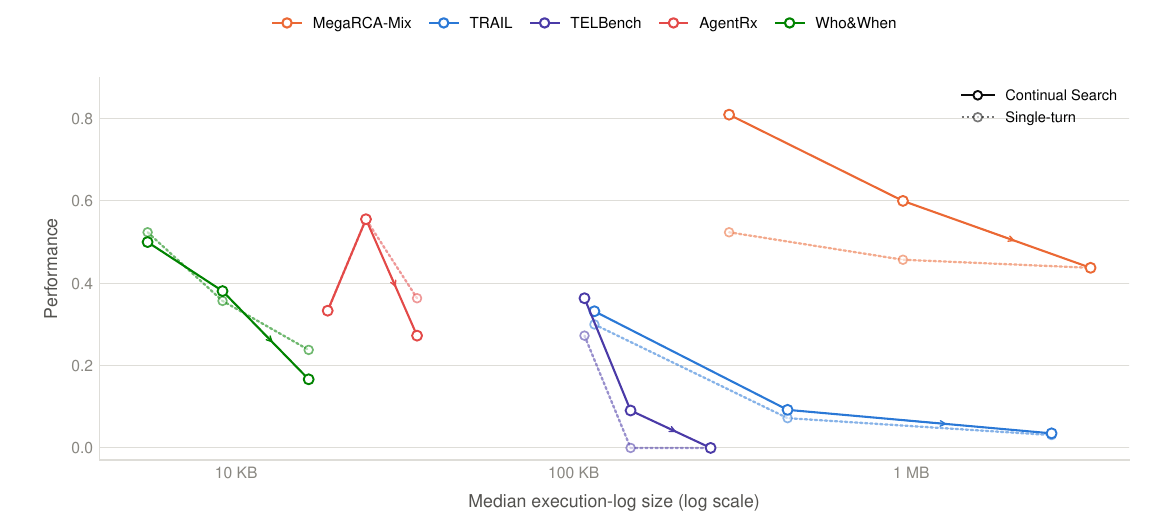}
\caption{Attribution performance of Opus-4.8 across evaluation log size terciles within each benchmark. Solid lines denote Continual Search, while dotted lines represent the corresponding single-turn baseline. Each data point is plotted at the median execution-log size of its tercile and evaluated using the native benchmark metric.}
\label{fig:t4-by-length}
\end{figure}

Among the three benchmarks with the largest execution records—MegaRCA-Mix, TRAIL, and TELBench—both the single-turn and the Continual Search attributions decline as execution records grow (Figure~\ref{fig:t4-by-length}). Continual Search stays above the initial attribution in every tercile except the largest on MegaRCA-Mix and TELBench, where the two coincide. In the highest tercile, the median log size reaches 3.40 MB on MegaRCA-Mix and 2.60 MB on TRAIL. In contrast, Continual Search does not exhibit a similar trend on AgentRx and Who\&When, which feature substantially shorter execution logs, with tercile medians ranging from 19--34 KB and 5--16 KB, respectively.
Together with the aggregate results in
Tables~\ref{tab:long-trajectory-performance} and
\ref{tab:short-trajectory-performance}, these findings indicate that
Continual Search is most useful when enough evidence remains available for
the judge to continue exploring.

\subsection{Evidence Acquisition Across Turns} \label{sec:dedup-observation} 
\begin{figure}[H]
\centering
\includegraphics[width=0.9\textwidth]{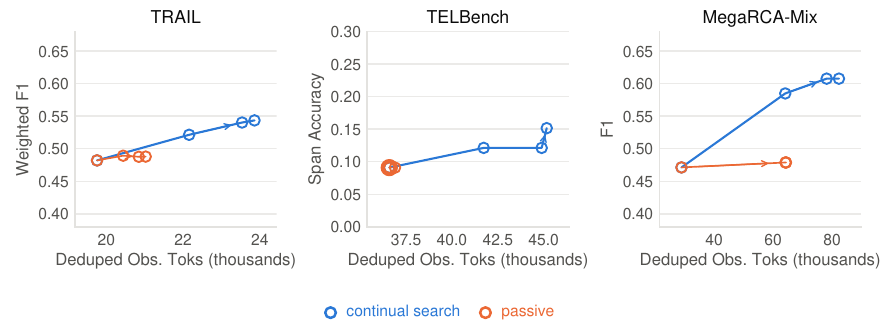}
\caption{Opus-4.8 attribution performance versus cumulative deduplicated observation tokens across continuation turns. Deduplicating repeated 13-gram spans ensures that horizontal movement strictly reflects the acquisition of novel evidence from the execution-log.}
\label{fig:dedup-observation}
\end{figure}

Continual Search is designed to extend the judge's search beyond the evidence examined in earlier turns. To quantify this, we track cumulative observation tokens after removing repeated 13-gram spans \citep{brown2020language} from the content returned by the judge's \texttt{read} tool. Repeated retrieval of the same material receives no additional credit, so increases in this measure reflect previously unseen evidence entering the judge's analysis. Figure~\ref{fig:dedup-observation} shows the same directional pattern for Opus-4.8 across MegaRCA-Mix, TRAIL, and TELBench. Continual Search extends farther into previously unseen evidence than Passive Continuation as attribution performance improves.

The absolute magnitude of new evidence acquired scales naturally with the size of the underlying execution records. The separation between the two methods is most pronounced on the large trajectories of MegaRCA-Mix, where Passive Continuation rapidly saturates while Continual Search maintains a steady intake of unseen evidence. TRAIL and TELBench exhibit the same underlying dynamic on a smaller scale. Across all three long-trajectory benchmarks, the performance gains yielded by Continual Search coincide with a broader, more exhaustive search over the available execution record.

Additional ablations on artifact-level evidence coverage, and stability across independent judge runs are presented in Appendix~\ref{sec:appendix-additional-ablations}.

% \section{Applications} \label{sec:applications}
% Improved root-cause attribution is valuable when it informs downstream
% diagnostic decisions. We evaluate whether a general failure-attribution
% system augmented with Continual Search can identify benchmark defects
% competitively with a purpose-built auditing system.

\section{Using Continual Search for Benchmark Diagnosis} \label{sec:aba} 
The interaction-centric taxonomy of \citet{raj2026modelharness} includes four failure modes specifically associated with benchmark-side defects: \emph{Instruction-Grader Mismatch}, \emph{Service Failure}, \emph{Stale State Delivery}, and \emph{Mistranslation}. We test whether Continual Search enables a general-purpose classifier using this taxonomy to reliably flag benchmark errors. As a specialized baseline, we compare against Automated Benchmark Analysis (ABA) \citep{aba}, a purpose-built auditing framework that evaluates tasks using a severity-scored rubric. We evaluate both methods using Opus-4.8 on the 50 MegaRCA-Mix trials. To ensure a strictly controlled evidence base, we bypass ABA's standard collector agent and provide both the ABA pipeline and the taxonomy classifier with identical access to the complete MegaRCA-Mix evaluation records, including trajectories, verifier outputs, and environment artifacts. 

\begin{figure}[H] 
\centering 
\begin{minipage}[c]{0.51\textwidth} 
  \centering 
  \includegraphics[width=\textwidth]{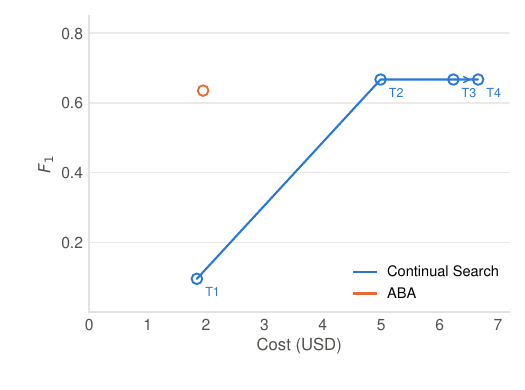} 
\end{minipage} 
\hfill 
\begin{minipage}[c]{0.48\textwidth} 
  \centering
  \small
  \rowcolors{2}{tableStripe}{white}
  \begin{tabular}{@{}lccc@{}} 
    \toprule 
    & Precision & Recall & $F_1$ \\ 
    \midrule 
    T1 & \textbf{1.00} & 0.05 & 0.10 \\ 
    T2 & \textbf{1.00} & 0.50 & \textbf{0.67} \\ 
    T3 & \textbf{1.00} & 0.50 & \textbf{0.67} \\ 
    T4 & \textbf{1.00} & 0.50 & \textbf{0.67} \\ 
    \midrule 
    ABA & 0.51 & \textbf{0.83} & 0.63 \\ 
    \bottomrule 
  \end{tabular} 
\end{minipage} 
\caption{\textbf{Benchmark audit performance on MegaRCA-Mix using Opus-4.8.} Comparison of the taxonomy classifier under Continual Search against the specialized ABA baseline across Turns 1 through 4. \textbf{Left:} $F_1$ score plotted against mean trajectory cost, showing that Continual Search surpasses ABA by turn 2. \textbf{Right:} Detailed metric breakdown, demonstrating that performance gains are driven by a sharp rise in Recall across turns. The highest score per column is bolded.}
\label{fig:aba} 
\end{figure}

Continual Search turns a taxonomy classifier into a more reliable benchmark diagnostic tool than ABA. As detailed in Figure~\ref{fig:aba}, the initial single-turn attribution yields an $F_1$ score of just $0.10$, severely lagging behind ABA's $0.63$. A single continuation turn immediately bridges this gap, raising the classifier's $F_1$ to $0.67$, where it stays through turn 4. This leap is driven entirely by recall, which surges from $0.05$ at T1 to $0.50$ at T2, while precision stays at $1.00$. By actively retrieving overlooked environment artifacts across successive turns, Continual Search allows a broad failure taxonomy to identify subtle benchmark defects more effectively than a dedicated diagnostic pipeline, albeit at a higher inference cost.

\section{Limitations}
\label{sec:limitations}

While Continual Search improves root-cause attribution, several limitations bound our analysis. Primary among these is the partial visibility of agent execution trajectories. Many evaluated trajectories are generated by proprietary models that do not expose their thinking tokens. The true causal failure might happen during these unobserved steps and only become visible later in the execution. Therefore, our results should be interpreted as root-cause attribution over the observable record.

The low absolute performance of frontier models on these benchmarks highlights broader evaluation challenges. First, we suspect that current failure taxonomies might contain overlapping categories, which could artificially lower classification scores. Second, establishing reliable human ground truth over massive execution logs is inherently difficult and introduces potential label noise.

% A second limitation concerns what our observation measures can establish. Our
% claims rest on comparisons between conditions rather than on per-trial
% associations within a condition, and this distinction matters because the amount
% of evidence a judge acquires is not assigned by the experiment. It responds to
% the trial, and to the judge's own progress on it, since a judge that has not yet
% reached a satisfying explanation has more reason to keep searching. The size of
% an evaluation record compounds this, because it governs both how much
% unexamined material remains and how difficult the attribution is, a pairing
% already visible in the log-size analysis of Section~\ref{sec:t4-by-length}. Correlating
% the evidence a judge acquired with the improvement it achieved therefore does
% not isolate the contribution of the search itself in either direction, and we do
% not read our observation measures that way. Establishing which component of
% continued search produces the gain would require a condition that preserves the
% continuation and its reconsideration pressure while withholding access to
% evidence the judge has not already examined, which we leave to future work.

\bibliography{custom}

\clearpage
\appendix

\section{Additional Ablations}
\label{sec:appendix-additional-ablations}

\subsection{Artifact-Level Evidence Coverage}

\begin{figure}[H]
\centering
\includegraphics[width=0.5\textwidth]{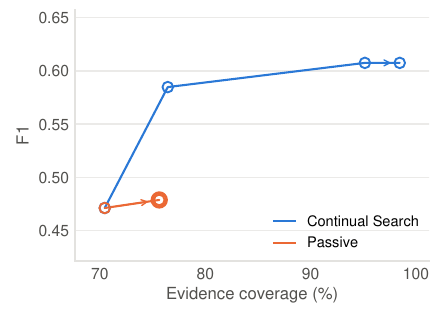}
\caption{\textbf{Attribution performance versus evidence coverage.} Mean $F_1$ score plotted against mean artifact coverage for Opus-4.8 on MegaRCA-Mix. Continual Search progressively inspects novel artifacts across successive turns to improve diagnostic performance, while Passive Continuation remains restricted to its initial evidence subset.}
\label{fig:coverage-vs-f1}
\end{figure}

To evaluate how search behavior impacts attribution performance, we measure the proportion of diagnostic evidence inspected by the judge across evaluation turns. By logging tool calls issued by the agentic judge, we track access to key artifacts, including agent trajectories, environment configurations, verifier outputs, and system logs. As illustrated in Figure~\ref{fig:coverage-vs-f1}, both search strategies start at an identical $70.5$\% evidence coverage at turn 1, reflecting their shared initial context. Continual Search subsequently expands evidence retrieval to $95.2$\% by turn 3 and $98.5$\% by turn 4. In contrast, Passive Continuation plateaus at $75.6$\% from turn 2 onward. This progressive acquisition of unread evidence directly drives diagnostic gains, demonstrating a strong positive correlation between artifact coverage and attribution performance.

% TODO: connection pass@4 vs turn4. turn4 is a better use of pass@k. leverage turn4 better than pass@4

Auxiliary artifacts account for most of this coverage gap. Although the judge inspects the primary agent trajectory in 38 of 50 trials by turn 2, secondary files such as system logs, environment configurations, and raw session records are typically accessed only in later turns. As shown in Figure~\ref{fig:artifact-coverage}, inspecting these previously unexamined files resolves initial misdiagnoses produced from trajectory data alone.

\begin{figure}[H]
\centering
\footnotesize GAIA2 \citep{froger2026gaia2}, adaptability config, \texttt{scenario\_universe\_21\_5e0gvz}
\par\vspace{3pt}
\begin{trajectoriespan}{scaleMediumGray}{What the user asked for}
``Cancel any appointments \ldots\ for this upcoming Saturday and schedule an
event titled `Film Production Day' \ldots\ adding my friend as an attendee.
Send him an email after scheduling \ldots\ If my friend says he can't make it,
accept any suggested date and time he proposes and reschedule \ldots\ If he
can't make it and doesn't suggest another date or time, cancel the Film
Production Day you scheduled.''
\end{trajectoriespan}
\par\vspace{7pt}
\footnotesize
\begin{trajectoriespan}[colback=red!3!white]{caseGray}{Turn 2 --- agent trajectory only}
``\ldots the trigger is a ConditionCheckEvent gated on the phase-1
\texttt{send\_message\_to\_user} count \ldots No positive evidence of a broken
environment exists, so \textbf{Service Failure / Stale State Delivery} fail the
`non-model only when a broken-env fact is established' bar. \ldots one
perfunctory 180s wait (returning in ${\sim}$3 real sec).''
\par\vspace{3pt}
$\rightarrow$ \textbf{owner --- model :: Satisficing}\quad\textbf{\textcolor{red!55!black}{\ding{55}}}
\end{trajectoriespan}
\par\vspace{5pt}
\begin{trajectoriespan}[colback=green!4!white]{contSearchBlue}{Turn 3 --- opens \texttt{config.json} and the agent console log, both previously unread}
{\raggedright
\texttt{are\_wait\_for\_notification(180)} $\rightarrow$
\texttt{\{"notifications": [], "waited\_seconds": 180, "iterations": 1\}}
\par\vspace{2pt}
\texttt{config.json}: \texttt{environment.env = \{\}}
\par}
\vspace{4pt}
``This is decisive new evidence, and it corrects a factual premise my prior
label rested on. \ldots My turns 1--2 rested on `the wait returned in
${\sim}$3 real seconds; the agent gave the env no chance' --- a \textbf{factual
error} about the sim-time semantics. The agent did \emph{not} stop at a low bar:
it waited the complete due-window and the environment produced nothing.''
\par\vspace{3pt}
$\rightarrow$ \textbf{env (external) --- model :: Stale State Delivery}\quad\textbf{\textcolor{green!45!black}{\ding{51}}}
\end{trajectoriespan}
\caption{Qualitative example from GAIA2 demonstrating how Continual Search uses non-trajectory artifacts to resolve misattributions. At turn 2, the judge relies solely on the agent trajectory, concludes that the agent terminated early, and assigns the failure to model \emph{Satisficing}. At turn 3, the judge inspects the previously unread \texttt{config.json} file and console log. These records reveal that the environment failed to respond during the full wait window, correcting the initial assumption to \emph{Stale State Delivery}. Excerpts are quoted verbatim.} \label{fig:artifact-coverage} \end{figure}

\subsection{Stability Across Independent Judge Runs} \label{sec:appendix-trail-seed-stability} 

The results reported in Table~\ref{tab:ensemble-judges} reflect a single judge run. To account for the inherent stochasticity of LLM outputs \citep{zhao2026jaggedjudges,dutta2026stability}, we test whether the advantage of Continual Search over Passive Continuation persists across independent runs. Specifically, we execute the Opus-4.8 evaluation on TRAIL three additional times, independently resampling the judge from the initial turn onward.

\begin{table}[H] 
\centering 
\small 
\rowcolors{3}{white}{tableStripe}
\begin{tabular}{@{}l ccc ccc@{}} 
\toprule
& \multicolumn{3}{c}{Joint Accuracy} & \multicolumn{3}{c}{Weighted F1} \\
\cmidrule(lr){2-4}\cmidrule(lr){5-7}
Run & Single-turn & Passive & Continual Search & Single-turn & Passive & Continual Search \\ 
\midrule
1 & 0.133 & 0.135 & \textbf{0.152} & 0.482 & 0.488 & \textbf{0.543} \\
2 & 0.152 & 0.154 & \textbf{0.197} & 0.465 & 0.484 & \textbf{0.505} \\
3 & 0.129 & 0.135 & \textbf{0.189} & 0.439 & 0.452 & \textbf{0.516} \\
4 & 0.138 & 0.145 & \textbf{0.182} & 0.475 & 0.484 & \textbf{0.531} \\
\bottomrule 
\end{tabular} 
\caption{Performance stability across four independent Opus-4.8 judge runs on TRAIL with $n=148$. Each run is independently sampled from the first turn onward, and the highest score per metric is bolded.} 
\label{tab:trail-seed-stability} 
\end{table}

Continual Search outperforms Passive Continuation across all four runs under both metrics. Absolute gains over Passive Continuation range between 0.037 and 0.055 for Joint Accuracy, and between 0.021 and 0.064 for Weighted F1. Although single-turn baseline accuracy fluctuates slightly across runs from 0.129 to 0.153, the relative performance advantage of Continual Search remains positive and consistent.

\section{Judge Configuration and Prompts}
\label{sec:appendix}

\subsection{Judge Configuration}
\label{sec:appendix-judge-config}

We use GPT-5.5 and Opus-4.8 as our two primary judges. GPT-5.5 is run with reasoning effort set to \texttt{xhigh}, while Opus-4.8 uses adaptive thinking with \texttt{max} effort. Both judges are invoked through the Claude Agent SDK. The judge panel in Table~\ref{tab:ensemble-judges} additionally includes Opus-4.7, configured with adaptive thinking and \texttt{max} effort, and Grok-4.6, with reasoning effort at xAI's own default of \texttt{high}. The judges have access to \texttt{Bash}, \texttt{Grep}, \texttt{Glob}, \texttt{Write}, \texttt{Edit}, and \texttt{Task}.

\raggedbottom
\subsection{Judge Prompts}
\label{sec:appendix-prompts}

This section gives every judge prompt verbatim. Braces and angle brackets mark the values that the runner fills in for each task.

\subsubsection{Turn 1 Prompts}
\label{sec:appendix-turn1}

Each benchmark keeps its published turn-1 prompt, except MegaRCA-Mix, whose prompt is introduced in this work. Because the execution records are long, every prompt gives the record as a file on disk instead of inlining it.

\paragraph{MegaRCA-Mix.} The prompt opens with a governing rule that applies to every turn of the session.

\begin{stickynote}{MEGARCA-MIX \,\textperiodcentered\, TURN 1}
\begin{promptverb}
GOVERNING RULE — applies to everything in this session, every turn. The
taxonomy is rooted in causation, not symptom: "find the FIRST failure — the terminal
error is only a symptom" ({judge_dir}/judge_prompt.txt). Every label you assign must be
anchored to the first point where the trajectory diverges from what an ideal,
well-exploring, cue-sensitive agent would have done — never to how the trial ended up.

STANDING DIRECTIVE — apply this at every step, not just once. Before accepting any
conclusion, pause and critique it as a skeptical reviewer would: does the evidence
survive scrutiny, or are you pattern-matching to what looks plausible? If you can poke a
hole in it, redo it before moving on.

The fault isn't always the model's. When a condition is true in fact — a broken env, a
wrong grader, a bad tool, a dropped instruction — it stays a non-model fault even if the
agent also stumbled; don't reassign it to the model just because the recovery was clumsy.

You are the FINAL adjudicating judge for ONE trial: `{trial_id}` of task `{task_id}`.
This is turn 1 of 2 — derive this trial's root cause, informed by the phase-1 votes but
not deferring to them.

LOGGING: keep your OWN log file inside {log_dir} (create it if needed). It's a live
scratchpad — append as you go, note anything you had to skip or defer.

Gather:
  1. the raw trial input: {traj_file} — peek into the referenced files rigorously,
     without being lazy.
  2. the phase-1 votes for THIS trial:
{votes_block}

You are adjudicating this trial ALONE. Do not look for, or reason from, sibling trials of
the same task — this judgement must rest on this trial's own trajectory.

Find this trial's first-divergence point from the raw trajectory. Use the votes as one
input, and decide the label from the trajectory evidence. Assign
(edge, fault_side, failure_mode) from {judge_dir}/taxonomy.py with a quoted
first-divergence citation.

Do not write any output file yet — hold your result in this conversation; turn 2 builds
on it.
\end{promptverb}
\end{stickynote}

\paragraph{TRAIL.} We use the published error-localization prompt of \citet{deshpande2025trail} and give the trace as a file on disk.

\begin{stickynote}{TRAIL \,\textperiodcentered\, TURN 1}
\begin{promptverb}
Follow the taxonomy below carefully follow the instructions and provide the output in the same format as the example.

# Taxonomy
├── Reasoning Errors
│   ├── Hallucinations
│   │   ├── Language-only
│   │   └── Tool-related (fabricating tool outputs/capabilities)
│   ├── Information Processing
│   │   ├── Poor Information Retrieval (Tried to find information that was not relevant to the task)
│   │   └── Tool Output Misinterpretation (Made assumptions about the tool output or used the tool output in an incorrect context)
│   ├── Decision Making
│   │   ├── Incorrect Problem Identification (Misunderstood the overall task or the local task)
│   │   ├── Tool Selection Errors (Used the wrong tool for the task)
│   └── Output Generation
│       ├── Formatting Errors (Errors with formatting and execution of code or structuring of output in a specific format)
│       └── Instruction Non-compliance (Failed to perform the task provided and instead did something else)
├── System Execution Errors
│   ├── Configuration
│   │   ├── Tool Definition Issues (The tool was not defined correctly by the user or contains some errors that make it inconsistent with its description. For example, web search tool was defined as a calculator tool)
│   │   └── Environment Setup Errors (includes permission problems and inability to access resources or API keys)
│   ├── API Issues
│   │   ├── Rate Limiting (Like 429)
│   │   ├── Authentication Errors (Like 401/403)
│   │   ├── Service Errors (Like 500)
│   │   └── Resource Not Found (Like 404)
│   └── Resource Management
│       ├── Resource Exhaustion (includes memory overflow)
│       └── Timeout Issues (The system took too long to respond)
├── Planning and Coordination Errors
│    ├── Context Management
│    │   ├── Context Handling Failures (includes window overflow and state tracking or forgetting important context)
│    │   └── Resource Abuse (Called the tool excessively due to memory issues)
│    └── Task Management
│        ├── Goal Deviation (The system deviated from the task or the subtask)
│        └── Task Orchestration (includes subtask coordination between agents and progress monitoring)

- Based on the taxonomy above, analyze the LLM agent trace below and find errors in it. 
- You must be exhaustive and find all the errors in the trace. Only include the final subcategories of the taxonomy (i.e. "Resource Not Found" and not "API Issues" or "System Execution Errors").
- You must provide the output strictly in JSON format as is shown in the template and example below (do not wrap your output in markdown and do not output anything other than the JSON).

Template for output:

{
    "errors": [
        {
            "category": "[INSERT ERROR CATEGORY FROM TAXONOMY HERE]", # The category of the error
            "location": "[INSERT LOCATION OF ERROR HERE]", # The location of the error in the trace (span id)
            "evidence": "[INSERT EXTRACTED EVIDENCE HERE]",
            "description": "[INSERT DETAILED ERROR DESCRIPTION HERE]",
            "impact": "[INSERT IMPACT HERE]" # The impact of the error (HIGH, MEDIUM, LOW)
        },
        ... # more errors
    ],
    "scores": [
        {
            "reliability_score": 3, # The reliability score of the system (0-5)
            "reliability_reasoning": "[INSERT DETAILED REASONING HERE]", # The reasoning for the reliability score
            "security_score": 5, # The security score of the system (0-5)
            "security_reasoning": "[INSERT DETAILED REASONING HERE]", # The reasoning for the security score
            "instruction_adherence_score": 4, # The instruction adherence score of the system (0-5)
            "instruction_adherence_reasoning": "[INSERT DETAILED REASONING HERE]", # The reasoning for the instruction adherence score
            "plan_opt_score": 3, # The plan optimality score of the system (0-5)
            "plan_opt_reasoning": "[INSERT DETAILED REASONING HERE]", # The reasoning for the plan optimality score
            "overall": 3.75 # The overall score of the system (0-5)
        }
    ]
}

Example output:

{
    "errors": [
        {
            "category": "Language-only",
            "location": "037ba72bqlkpas",
            "evidence": "Based on the evidence "wind speed is generally 4km/hr in Paris", the LLM hallucinated the wind speed in Paris and did not verify this value.",
            "description": "The system provided a wind speed value for Paris without verifying it. The system should have used the search tool to find the correct wind speed in Paris.",
            "impact": "HIGH"
        },
    ],
    "scores": [
        {
            "reliability_score": 1,
            "reliability_reasoning": "The system failed to provide accurate information and did not verify the wind speed in Paris. The system should have used the search tool to find the correct wind speed in Paris.",
            "security_score": 5,
            "security_reasoning": "No security issues were detected. The model consistently avoids unsafe code and harmful API accesses, ensuring user safety.",
            "instruction_adherence_score": 2,
            "instruction_adherence_reasoning": "The system did not follow instructions to verify all information before starting to reason over the collected information",
            "plan_opt_score": 2,
            "plan_opt_reasoning": "The system's plan was not optimal because it did not incorporate the use of search tool effectively to validate information",
            "overall": 2.5
        }
    ]
}

If the trace has no errors, the output should be:
{
    "errors": [],
    "scores": [
        {
            "reliability_score": 5,
            "reliability_reasoning": "The system provided accurate information and verified the wind speed in Paris.",
            "security_score": 5,
            "security_reasoning": "No security issues were detected. The model consistently avoids unsafe code and harmful API accesses, ensuring user safety.",
            "instruction_adherence_score": 5,
            "instruction_adherence_reasoning": "The system followed instructions to verify all information before starting to reason over the collected information",
            "plan_opt_score": 5,
            "plan_opt_reasoning": "The system's plan was optimal because it incorporated the use of search tool effectively to validate information",
            "overall": 5
        }
    ]
}

The data to analyze is as follows:

The trace is an OpenTelemetry JSON file on disk at:
    <trace-file path>

It has <number of spans> spans and is <file size> MB, which is too large to inline here -- read it from disk with your tools (Read, Grep, Bash) before answering. Notes on its shape:
  - Execution order is NESTED under `child_spans`, not a flat sequence.
  - Each span's id is the `span_id` field; that is what `location` must cite.
  - For a large file, prefer `python3 -c` with json to walk the span tree, or Grep for specific fields, over a single whole-file Read.
Examine every span before you answer -- the instruction above to be exhaustive applies to the whole trace, not just the parts that are easy to reach.

- Ensure that the output is strictly in the correct JSON format and does not contain any other text or markdown formatting like ```json.
- Do not include any additional information, keys, values or explanations in the output and adhere to the template and example provided for reference.
- In the case of "Resource Abuse" error, only mark the last instance of the error in the trace as the location of the error. For all other errors, you must mark the first instance of the error in the trace as the location of the error.
\end{promptverb}
\end{stickynote}

\paragraph{TELBench.} We use DRIFT's published "bare" error-span prompt \citep{wang2026drift} and give the spans as files on disk, since a single trajectory can reach 161K tokens.

\begin{stickynote}{TELBENCH \,\textperiodcentered\, TURN 1}
\begin{promptverb}
You are evaluating one deep-research trajectory for span-level error localization.
This is a bare single-call evaluation: read the full question and all ordered spans once, then predict final error spans directly.

Mark a span only if the span itself contains a committed harmful mistake, an unsupported committed conclusion, a harmful premature finalization, or a harmful continuation.
Do not mark harmless exploration, ordinary evidence gaps, isolated tool failures without commitment, retries, search queries, tentative candidate pivots, or generic uncertainty.
Prefer a sparse set of committed harmful spans. If the actual harmful commitment appears only in the final report, output only that final span.
If an early span already commits to the wrong answer path or harmful no-answer decision, mark that earliest committed span and any later spans that explicitly rely on or finalize it.
If there is no committed harmful error, return an empty error_span_ids list.

Return JSON only. Do not include markdown or explanations outside JSON.
Schema: {"traj_id": "<traj-id>", "error_span_ids": ["s004", "s007"], "earliest_harmful_span_id": "s004", "reasons": [{"span_id": "s004", "reason": "short string"}]}

{
  "question": "<question text>",
  "traj_id": "<traj-id>",
  "n_spans": <number of spans>,
  "spans_index": "<spans index path>",
  "spans_dir": "<spans directory>",
  "spans": "The ordered spans are NOT inlined here -- this trajectory is too large to inline. They are on disk. Read `spans_index` first: it lists every span_id in trajectory order with its size and a short preview. Then read individual spans in full from `spans_dir`/<span_id>.txt. Read every span before answering; a span you have not opened is a span you cannot rule in or out."
}
\end{promptverb}
\end{stickynote}

\paragraph{AgentRx.} We keep the published taxonomy and checklist of AgentRx \citep{barke2026agentrx} and ask the judge to list every candidate failure, which the later turns narrow down.

\begin{stickynote}{AGENTRX \,\textperiodcentered\, TURN 1}
\begin{promptverb}
GIVEN INPUT:
- a full trajectory of an agent's interaction with a user (step-indexed)
- the ground-truth tool-call/action sequence the agent should have made
- optional: expected responses/outputs for some steps

YOUR TASK is to determine why the agent failed, which failure category applies from the taxonomy below, and exactly which step index the failure occurred at.

FAILURE TAXONOMY CATEGORIES:
The failure taxonomy has the following categories:

1. Instruction/Plan Adherence Failure: Goal is correct, but the agent deviates from the required plan by ignoring directives and skipping steps despite having enough information. 
This covers both under-execution (missed steps) and over-execution (unplanned or unnecessary actions, e.g., extra tool calls) that deviate from the static plan, domain policy or orchestrator plan.
   Checklist:
   - Can you state the user's goal, and do the agent's intent and end goal match that goal (i.e., the agent is not solving the wrong problem)?
   - Was all the required information already available at this step (user intent, required context, prior tool outputs)?
   - Is there a step where the ground-truth/policy requires an action (tool call, question, confirmation, ordering) and the agent did something different (skipped it / reordered it / added extra unneeded action)?

2. Invention of New Information: The agent introduces, removes, or alters information that is not grounded in any available input, context, or tool output. This includes fabricating unsupported facts, hallucinating details, or omitting relevant information.
   Checklist:
   - Can you pinpoint the exact invented/altered/omitted claim, value, or assumption the agent used?
   - Is that claim absent from all evidence available up to that step (user text, provided context, tool outputs)?
   - Did the agent rely on that claim to decide an action or produce the failing conclusion (not just harmless wording)?

3. Invalid Invocation: Tool call fails because the request is ill-formed (missing args, wrong fields/types, malformed query, schema mismatch).
   Checklist:
   - At the failure step, did the agent attempt a tool call with a concrete invocation payload/arguments?
   - Does the tool/runtime explicitly report a parse/validation/schema/syntax error for that call (e.g., missing field, invalid type, cannot parse, malformed query)?
   - Is the error NOT primarily a network/timeout/service-unavailable/endpoint-unreachable issue (infra/connectivity)?
   - Is the error NOT primarily a CAPTCHA/login/paywall refusal (access/guardrail block)?

4. Misinterpretation of Tool Output / Handoff Failure: The agent incorrectly reasons about its own or another agent's tool output, leading to incorrect assumptions or actions. This also includes cases where the agent considered only partial tool output.
   Checklist:
   - Before (or at) the failure step, did the agent receive tool output or handoff output that is relevant to the failing decision?
   - Did the agent state or imply a specific reasoning derived from that tool output?
   - Does that reasoning contradict the tool output, omit a crucial part, or reflect a clear computation/logic error relative to the output?

5. Intent-Plan Misalignment: Agent misunderstands the user's intent/constraints and pursues the wrong objective or violates key constraints due to misunderstanding.
   Checklist:
   - Do the agent's actions/plan optimize for a different goal OR violate a key constraint (not a minor wording/format issue)?
   - Is the misalignment due to misunderstanding of intent/constraints (rather than missing required info from the user/context/tool outputs)?
   - Is the misalignment not primarily caused by a tool error (invalid invocation, infra failure, or access/guardrail block)?

6. Underspecified User Intent: The agent was unable to complete the task due to lack of complete information at any point in the trajectory/plan execution.
   Checklist:
   - Can you identify a specific missing piece of information that is required to proceed correctly (e.g., date, address, account id, item variant)?
   - Is that information absent from all evidence available up to that step (user text, provided context, and tool outputs)?
   - Did the agent fail because it proceeded without obtaining this information OR because it did not ask for it when needed?

7. Intent Not Supported: Requested action cannot be performed with available tools/capabilities.
   Checklist:
   - Is the user requesting an action that requires an external capability/tool (e.g., listen to audio, access a private system, perform a human action)?
   - Given the tool set available in this environment, is there no tool that can perform the requested action?
   - Is the failure not primarily caused by infrastructure/connectivity issues?

8. Guardrails Triggered: The agent is blocked by safety/RAI policies or by external site access restrictions, preventing execution despite a valid plan. 
   Checklist:
   - Is there an explicit refusal/block signal (policy refusal, CAPTCHA, login required, 403, paywall, robots.txt, automation forbidden)?
   - Would the plan be feasible and correct if this block were removed (i.e., the agent is not pursuing the wrong goal/constraints)?
   - Is the failure not primarily due to malformed tool invocation (schema/syntax/args validation error)?
   - Is the failure not primarily due to infrastructure/connectivity issues (timeouts, endpoint unreachable)?

9. System Failure: The agent faces a system connectivity issue while calling a particular tool like an endpoint not being reachable.
   Checklist:
   - At the failure step, did the agent attempt a tool call or rely on a tool that should have been callable?
   - Is there an explicit infra/connectivity error signal (timeout, connection refused, DNS failure, endpoint unreachable, service unavailable, premature termination)?
   - Is the failure not primarily a parse/validation/schema/syntax error caused by malformed arguments?

10. Inconclusive (USE SPARINGLY): None of 1-10 clearly apply; must provide a custom category label.
   Checklist:
   - If labeling as 10, did you provide a non-empty custom_category describing the failure type?



ROOT-CAUSE DETECTION ALGORITHM:

Step 1 — Locate the first failure: Scan the trajectory step-by-step from the start and record the first failure.
Step 2 — Check if that failure was resolved: Look ahead in the trajectory for evidence that the error was resolved. If yes → Resolved; if no such evidence → Not resolved.
Step 3 — Decide and continue:
If Resolved: continue scanning from the next step to find the next new failure, then repeat Step 2 for it.
If Not resolved: treat this step as the root-cause failure for the run and assign the taxonomy at this step.

Output a JSON object in the following format:
{
    "taxonomy_checklist_reasoning": <string>,
    "reason_for_failure": <string>,
    "failure_case": <int 1-10>,
    "reason_for_index": <string>,
    "index": <int>
}

The trajectory is NOT included above -- it is a JSON file. Read it in full with the Read tool before answering: it is a list of entries in file order, each with an explicit "step": N field. Use THAT field's value -- never the file's line number.

Your task on this turn is ENUMERATION, not selection. List EVERY candidate failure across the WHOLE trajectory -- do not focus on just one part or stop at the first one you find. Exhaustive means exhaustive: check the beginning, middle, and end equally; a subtle early issue is as valid a candidate as an obvious late one. For each candidate, give the step index, the taxonomy category, a verbatim quote of the bad commitment/action, and a verbatim quote of what it contradicts (a policy line, an earlier tool output, or the user's stated intent). Do NOT judge yet whether it was resolved or whether it's the root cause -- leave those fields null. List at least 2 candidates if the trajectory has any plausible ambiguity at all.

Output a JSON object in the following format:
{
    "candidates": [
        {"step": <int>, "category": <taxonomy category name>,
         "commitment_quote": <verbatim quote of the bad action/claim>,
         "violated_quote": <verbatim quote of what it contradicts -- policy, prior tool output, or user intent>,
         "resolved": <true|false|null>, "resolved_evidence": <string, quote or "">,
         "terminal_connected": <true|false|null>, "terminal_evidence": <string, quote or "">}
    ],
    "index": <int, your current best guess, for tracking only>,
    "failure_case": <int 1-10, your current best guess, for tracking only>
}

Trajectory file: <trajectory-file path>

Read it, then output the JSON.
\end{promptverb}
\end{stickynote}

\paragraph{Who\&When.} We use the published "all\_at\_once" template of Who\&When \citep{zhang2025whowhen} and give the conversation as a file on disk. The first line is the system prompt.

\begin{stickynote}{WHO\&WHEN \,\textperiodcentered\, TURN 1}
\begin{promptverb}
You are a helpful assistant skilled in analyzing conversations.

You are an AI assistant tasked with analyzing a multi-agent conversation history when solving a real world problem. The problem is:  <problem statement>
The Answer for the problem is: <gold answer>
Identify which agent made an error, at which step, and explain the reason for the error. Here's the conversation:

It is not inlined here. The conversation is a JSON file on disk at:
    <conversation-file path>

Read it in full with the Read tool before answering. It is a list of <number of steps> entries in file order, each with an explicit "step": N field and the speaking agent's name in "agent". Use THAT "step" field's value as the step number -- never the file's line number. An entry whose content is long is split into a "content_chunks" list; read every chunk of it. Examine every entry before you answer -- the first mistake can be early and quiet as easily as late and obvious.

Based on this conversation, please predict the following:
1. The name of the agent who made a mistake that should be directly responsible for the wrong solution to the real world problem. If there are no agents that make obvious mistakes, decide one single agent in your mind. Directly output the name of the Expert.
2. In which step the mistake agent first made mistake. For example, in a conversation structured as follows: 
            {
                "agent a": "xx",
                "agent b": "xxxx",
                "agent c": "xxxxx",
                "agent a": "xxxxxxx"
            },
            each entry represents a 'step' where an agent provides input. The 'x' symbolizes the speech of each agent. If the mistake is in agent c's speech, the step number is 2. If the second speech by 'agent a' contains the mistake, the step number is 3, and so on. Please determine the step number where the first mistake occurred.
3. The reason for your prediction.Please answer in the format: Agent Name: (Your prediction)
 Step Number: (Your prediction)
 Reason for Mistake: 
\end{promptverb}
\end{stickynote}

\subsubsection{Continuation Prompts}
\label{sec:appendix-cont}

\begin{figure}[t]
\centering
\includegraphics[width=0.92\textwidth]{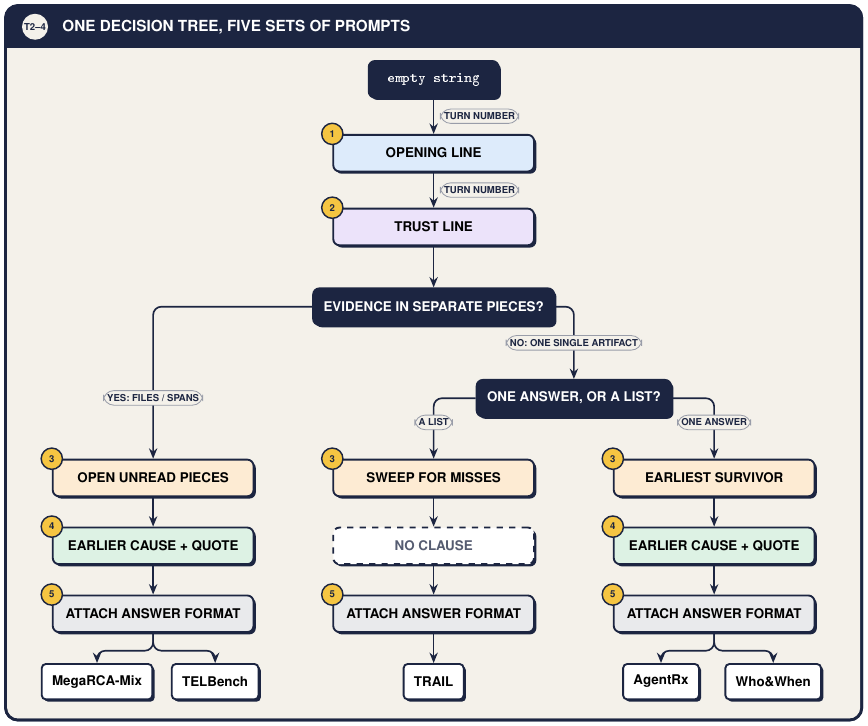}
\caption{The decision tree that assembles every Continual Search prompt for turns 2--4. Each prompt starts empty and gains one sentence at each step. The numbered badges mark the five parts of every prompt, namely the opening line (1), trust line (2), search action (3), change rule (4), and answer format (5). The full prompts are given below.}
\label{fig:prompt-tree}
\end{figure}

Turns 2--4 continue within the same session described in Section~\ref{sec:continual-search}, and each turn receives the full conversation history and a new instruction. All turns retain the benchmark's original output schema. To make Continual Search general, a single decision tree assembles the prompts for turns 2--4 from a fixed set of generic sentences (Figure~\ref{fig:prompt-tree}). The sentences were written once, without reference to any particular benchmark, and a benchmark enters only through the nouns that fill their slots. A new benchmark therefore needs no new wording, only the answers to the two questions in the tree, namely whether its evidence comes in separate pieces and whether its answer is one item or a list. Benchmarks on the same branch receive nearly identical prompts, which differ by 7\% on average in character-level string similarity.

The turn number selects the opening line and the trust line. The first question asks whether the evidence is stored as separate pieces that the judge can leave unopened. MegaRCA-Mix stores each run as separate files, and TELBench splits each trajectory into spans. For these two benchmarks, turn 2 asks the judge to derive its answer again from the start, and turns 3 and 4 ask it to open the unopened pieces. When the evidence is a single artifact that the judge has already read, the second question asks whether the answer is one item or a list, which follows the benchmark's metric. TRAIL scores a set of errors, so its turns ask for another sweep of the least-covered parts of the trace. AgentRx and Who\&When score a single step, so their turns test each candidate and keep the earliest one that was never fixed and still leads to the failure. At turns 3 and 4, the tree adds a change rule for benchmarks whose answer is a single item. The answer may then move only to an earlier cause that the judge supports with an exact quote. TRAIL instead asks for a list of errors. Its prompts, under both Continual Search and Passive Continuation, already tell the judge to keep earlier findings unless it can give a specific reason to drop them, so the tree adds no rule for it. Each prompt ends with the benchmark's own answer format.

Each prompt ends with the benchmark's answer format, which the notes below omit except for TRAIL, whose format is a single sentence at the end of each note.

\paragraph{MegaRCA-Mix and TELBench.} Highlighted text in the TELBench notes differs from MegaRCA-Mix. The runner replaces \texttt{\{unread\_block\}} with the list of unopened pieces.

\begin{tcbraster}[raster columns=2, raster equal height=rows, raster column skip=10pt, raster row skip=14pt, raster before skip=14pt, raster after skip=10pt]
\begin{stickycell}{MEGARCA-MIX \,\textperiodcentered\, TURN 2}
\begin{promptdiff}
This is pass 2 over the same trajectory.

Your turn-1 label came from a single first reading and has not been tested yet, so do not treat it as established.

Re-read the trajectory from the start and derive the label again, independently of what you concluded at turn 1. Then compare the two derivations and keep whichever is supported by the better evidence.
\end{promptdiff}
\end{stickycell}
\begin{stickycell}{TELBENCH \,\textperiodcentered\, TURN 2}
\begin{promptdiff}
This is pass 2 over the same trajectory.

Your turn-1 |pdiff[answer] came from a single first reading and has not been tested yet, so do not treat it as established.

Re-read the trajectory from the start and derive the |pdiff[answer] again, independently of what you concluded at turn 1. Then compare the two derivations and keep whichever is supported by the better evidence.
\end{promptdiff}
\end{stickycell}
\begin{stickycell}{MEGARCA-MIX \,\textperiodcentered\, TURN 3}
\begin{promptdiff}
This session continues beyond the number of turns announced at the start, and every instruction from turn 1 still holds.

Your current label is still provisional.

There are records of this run you have not opened, listed here:

{unread_block}

Open all of them in full before answering; unopened evidence cannot support any conclusion.

Your standing label may change only to a cause occurring earlier than the current one, supported by an exact quote. Without such a quote, restate it unchanged. Changing it merely to appear responsive is forbidden.
\end{promptdiff}
\end{stickycell}
\begin{stickycell}{TELBENCH \,\textperiodcentered\, TURN 3}
\begin{promptdiff}
|pdiff[This] |pdiff[is] |pdiff[pass] |pdiff[3] |pdiff[over] |pdiff[the] |pdiff[same] |pdiff[trajectory.]

Your current |pdiff[answer] is still provisional.

There are |pdiff[spans] you have not opened, listed here:

{unread_block}

Open all of them in full before answering; unopened evidence cannot support any conclusion.

Your standing |pdiff[answer] may change only to a cause occurring earlier than the current one, supported by an exact quote. Without such a quote, restate it unchanged. Changing it merely to appear responsive is forbidden.
\end{promptdiff}
\end{stickycell}
\begin{stickycell}{MEGARCA-MIX \,\textperiodcentered\, TURN 4}
\begin{promptdiff}
This is the final turn; nothing can change after it.

Your current label becomes final only after this turn's check.

Whatever remains unopened among the records of this run listed here must be opened in full before you finalize:

{unread_block}

Unopened evidence cannot support any conclusion.

Your standing label may change only to a cause occurring earlier than the current one, supported by an exact quote. Without such a quote, restate it unchanged. Changing it merely to appear responsive is forbidden.
\end{promptdiff}
\end{stickycell}
\begin{stickycell}{TELBENCH \,\textperiodcentered\, TURN 4}
\begin{promptdiff}
This is the final turn; nothing can change after it.

Your current |pdiff[answer] becomes final only after this turn's check.

Whatever remains unopened among the |pdiff[spans] listed here must be opened in full before you finalize:

{unread_block}

Unopened evidence cannot support any conclusion.

Your standing |pdiff[answer] may change only to a cause occurring earlier than the current one, supported by an exact quote. Without such a quote, restate it unchanged. Changing it merely to appear responsive is forbidden.
\end{promptdiff}
\end{stickycell}
\end{tcbraster}

\paragraph{TRAIL.} The last paragraph of each note is TRAIL's own instruction, which its runner appends in both conditions.

\begin{stickynote}{TRAIL \,\textperiodcentered\, TURN 2}
\begin{promptdiff}
This is pass 2 over the same trace.

Your turn-1 list of errors came from a single first reading and has not been tested yet, so do not treat it as established.

Re-read the trace, focusing on the parts you covered least on the first pass, and add any newly found errors to your list of errors.

Do not drop or restate earlier findings unless you find a specific, checkable reason one of them is wrong. Output the full updated list of errors, in the same JSON schema as before.
\end{promptdiff}
\end{stickynote}
\begin{stickynote}{TRAIL \,\textperiodcentered\, TURN 3}
\begin{promptdiff}
This is pass 3 over the same trace.

Your current list of errors is still provisional.

Go over the least-covered stretches of the trace again and add anything you find.

Do not drop or restate earlier findings unless you find a specific, checkable reason one of them is wrong. Output the full updated list of errors, in the same JSON schema as before.
\end{promptdiff}
\end{stickynote}
\begin{stickynote}{TRAIL \,\textperiodcentered\, TURN 4}
\begin{promptdiff}
This is the final turn; nothing can change after it.

Your current list of errors becomes final only after this turn's check.

Make one final sweep of the trace and add anything new before finishing.

Do not drop or restate earlier findings unless you find a specific, checkable reason one of them is wrong. Output your final, complete list of errors, in the same JSON schema as before.
\end{promptdiff}
\end{stickynote}

\paragraph{AgentRx and Who\&When.} Highlighted text in the Who\&When notes differs from AgentRx.

\begin{tcbraster}[raster columns=2, raster equal height=rows, raster column skip=10pt, raster row skip=14pt, raster before skip=14pt, raster after skip=10pt]
\begin{stickycell}{AGENTRX \,\textperiodcentered\, TURN 2}
\begin{promptdiff}
This is pass 2 over the same trajectory file.

Your turn-1 answer came from a single first reading and has not been tested yet, so do not treat it as established.

Re-read the trajectory file in full and evaluate every earlier candidate independently, each on its own evidence rather than relative to the others, as a possible first cause of the failure. Also note any candidate you missed at turn 1.
\end{promptdiff}
\end{stickycell}
\begin{stickycell}{WHO\&WHEN \,\textperiodcentered\, TURN 2}
\begin{promptdiff}
This is pass 2 over the same |pdiff[conversation.]

Your turn-1 |pdiff[attribution] came from a single first reading and has not been tested yet, so do not treat it as established.

Re-read the |pdiff[conversation] in full and evaluate every earlier candidate |pdiff[mistake] independently, each on its own evidence rather than relative to the others, as a possible first cause of the failure. Also note any candidate |pdiff[mistake] you missed at turn 1.
\end{promptdiff}
\end{stickycell}
\begin{stickycell}{AGENTRX \,\textperiodcentered\, TURN 3}
\begin{promptdiff}
This is pass 3 over the same trajectory file.

Your current answer is still provisional.

For each candidate, decide two things from the trajectory file, with quoted evidence for each: whether a later step actually fixed it, and whether an unbroken causal path still leads from it to the final failure.

Your standing answer may change only to a cause occurring earlier than the current one, supported by an exact quote. Without such a quote, restate it unchanged. Changing it merely to appear responsive is forbidden.
\end{promptdiff}
\end{stickycell}
\begin{stickycell}{WHO\&WHEN \,\textperiodcentered\, TURN 3}
\begin{promptdiff}
This is pass 3 over the same |pdiff[conversation.]

Your current |pdiff[attribution] is still provisional.

For each |pdiff[candidate] |pdiff[mistake,] decide two things from the |pdiff[conversation,] with quoted evidence for each: whether a later step actually fixed it, and whether an unbroken causal path still leads from it to the final failure.

Your standing |pdiff[attribution] may change only to a cause occurring earlier than the current one, supported by an exact quote. Without such a quote, restate it unchanged. Changing it merely to appear responsive is forbidden.
\end{promptdiff}
\end{stickycell}
\begin{stickycell}{AGENTRX \,\textperiodcentered\, TURN 4}
\begin{promptdiff}
This is the final turn; nothing can change after it.

Your current answer becomes final only after this turn's check.

Restrict attention to candidates that are unfixed and still causally linked to the final failure. Take the earliest of them, verify it once more against the trajectory file, and answer.

Your standing answer may change only to a cause occurring earlier than the current one, supported by an exact quote. Without such a quote, restate it unchanged. Changing it merely to appear responsive is forbidden.
\end{promptdiff}
\end{stickycell}
\begin{stickycell}{WHO\&WHEN \,\textperiodcentered\, TURN 4}
\begin{promptdiff}
This is the final turn; nothing can change after it.

Your current |pdiff[attribution] becomes final only after this turn's check.

Restrict attention to |pdiff[candidate] |pdiff[mistakes] that are unfixed and still causally linked to the final failure. Take the earliest of them, verify it once more against the |pdiff[conversation,] and answer.

Your standing |pdiff[attribution] may change only to a cause occurring earlier than the current one, supported by an exact quote. Without such a quote, restate it unchanged. Changing it merely to appear responsive is forbidden.
\end{promptdiff}
\end{stickycell}
\end{tcbraster}

\paragraph{Passive Continuation.} It sends one sentence at every turn for every benchmark, followed by the benchmark's answer format.

\begin{stickynote}{ALL BENCHMARKS \,\textperiodcentered\, TURNS 2--4, PASSIVE}
\begin{promptverb}
Reconsider and reverify your current answer.
\end{promptverb}
\end{stickynote}

\paragraph{MegaRCA-Mix answer format.} The answer format for turns 2--4 is introduced in this work. Changes to the initial prediction are recorded with \texttt{changed\_from\_turn1} at turn 2 and with \texttt{changed\_from\_prev} at turns 3 and 4.

\begin{stickynote}{MEGARCA-MIX \,\textperiodcentered\, ANSWER FORMAT, TURNS 2--4}
\begin{promptverb}
OUTPUT — use the Write tool to save this exact JSON to {out_path}:
  {"trial_id": "{trial_id}", "task_id": "{task_id}",
    "edge": "<edge verbatim from taxonomy.py>",
    "fault_side": "<model|owner|env|orchestrator|tool|grader|third party|...>",
    "failure_mode": "<failure_mode verbatim from taxonomy.py>",
    "first_divergence": "...", "evidence_quote": "...", "rationale": "...",
    "changed_from_turn1": true|false}
`edge` and `failure_mode` must be the exact strings from taxonomy.py — they are the
machine-joinable answer.
\end{promptverb}
\end{stickynote}

\subsection{Evaluation Metrics} \label{sec:appendix-metrics} 
To ensure fair comparison, we evaluate Continual Search using the native metrics and scoring logic established by each benchmark. The exact mathematical formulations are detailed below.

\paragraph{MegaRCA-Mix.} 
MegaRCA-Mix relies on a set-based $F_1$ score to account for failures that may fall under multiple valid root-cause categories. Let $\hat{C}_i$ denote the set of taxonomy clusters predicted for trial $i$, and let $\mathcal{G}_i$ represent the set of all acceptable gold cluster sets. For any nonempty sets $A$ and $B$, the overlap is:
\begin{equation} 
F_1(A,B) = \frac{2|A \cap B|}{|A|+|B|}. 
\end{equation} 
A prediction is considered correct if it matches any of the acceptable gold sets, so the trial score is calculated as the maximum $F_1$ across $\mathcal{G}_i$. The final score averages these maximums across all $N$ trials:
\begin{equation} 
\mathrm{F1}_i = \max_{C \in \mathcal{G}_i} F_1(\hat{C}_i,C), \qquad \mathrm{F1} = \frac{1}{N}\sum_{i=1}^{N}\mathrm{F1}_i. 
\end{equation} 

\paragraph{TRAIL.} 
TRAIL evaluates both the location and the category of an error \citep{deshpande2025trail}. Let $\hat{E}_i$ and $E_i^\star$ denote the predicted and gold sets of location--category pairs for trial $i$. Joint Accuracy measures the recall of gold pairs:
\begin{equation} 
J_i = \begin{cases} 
\dfrac{|\hat{E}_i \cap E_i^\star|}{|E_i^\star|}, & |E_i^\star|>0, \\[5pt] 
0, & |E_i^\star|=0, 
\end{cases} 
\qquad \mathrm{JointAcc} = \frac{1}{N}\sum_{i=1}^{N} J_i . 
\end{equation} 
Joint Accuracy acts strictly as a recall metric, meaning it does not penalize the model for predicting additional, incorrect location--category pairs. We therefore also report TRAIL's Weighted F1 over error categories. Let $y_{ic} \in \{0,1\}$ and $\hat{y}_{ic} \in \{0,1\}$ indicate whether category $c$ appears in the gold and predicted sets, respectively. For each category:
\begin{equation} 
\begin{aligned} 
\mathrm{TP}_c &= \sum_i y_{ic}\hat{y}_{ic}, & \mathrm{FP}_c &= \sum_i (1-y_{ic})\hat{y}_{ic}, & \mathrm{FN}_c &= \sum_i y_{ic}(1-\hat{y}_{ic}), \\ 
F_{1,c} &= \frac{2\mathrm{TP}_c} {2\mathrm{TP}_c+\mathrm{FP}_c+\mathrm{FN}_c}. 
\end{aligned} 
\end{equation} 
The final Weighted F1 score averages these category-level scores, weighted by their true support $n_c=\sum_i y_{ic}$:
\begin{equation} 
\mathrm{WeightedF1} = \frac{\sum_c n_c F_{1,c}}{\sum_c n_c}. 
\end{equation} 

\paragraph{TELBench.} 
TELBench provides multiple span-level metrics, from which we report First-Error Accuracy (FEA) \citep{wang2026drift}. Let $\hat{s}_i^{(1)}$ and $s_i^{\star(1)}$ denote the first predicted and gold error-span IDs for trial $i$. The metric requires an exact match:
\begin{equation} 
\mathrm{FEA} = \frac{1}{N}\sum_{i=1}^{N} \mathbb{1}\!\left[ \hat{s}_i^{(1)} = s_i^{\star(1)} \right]. 
\end{equation} 

\paragraph{AgentRx and Who\&When.} 
Both AgentRx (Critical Step-index Accuracy) and Who\&When (Step-Level Accuracy) require the judge to exactly identify the integer index of the decisive failure step \citep{barke2026agentrx,zhang2025whowhen}. Let $\hat{k}_i$ and $k_i^\star$ denote the predicted and gold step indices:
\begin{equation} 
\mathrm{StepAcc} = \frac{1}{N}\sum_{i=1}^{N} \mathbb{1}\!\left[\hat{k}_i=k_i^\star\right]. 
\end{equation}

\end{document}